%% file: main.tex
\documentclass{applepaper}
\usepackage{soul}
\usepackage{tabularx}
\graphicspath{{figures/}}

\ifdefined\pdfsuppressptexinfo
\fi

\newcommand{\method}{MorphHOI\xspace}
\newcommand{\off}{\cellcolor{black!10}}

\hypersetup{
  pdftitle={One Policy, Many Bodies: Cooperative Human-Object Interaction with Heterogeneous Teams},
  pdfauthor={Zhihao Cao, Tianxu An, Chenhao Li, Chenyangguang Zhang, Stelian Coros, Marco Hutter},
}

\kicker{Research Article}
\title{One Policy, Many Bodies}
\subtitle{Cooperative Human-Object Interaction with Heterogeneous Teams}
\shorttitle{Cooperative Human-Object Interaction with Heterogeneous Teams}
\author{Zhihao Cao\affil{1}\quad
        Tianxu An\affil{1,2,\textdagger}\quad
        Chenhao Li\affil{1,3}\quad
        Chenyangguang Zhang\affil{1,4}\quad
        Stelian Coros\affil{2}\quad
        Marco Hutter\affil{1}}
\affiliations{\affiliation{1}{Robotic Systems Lab, ETH Zurich}\linksep
              \affiliation{2}{Computational Robotics Lab, ETH Zurich}\linksep
              \affiliation{3}{ETH AI Center}\linksep
              \affiliation{4}{Computer Vision and Geometry Group, ETH Zurich}}
\links{\href{https://morphhoi.github.io}{https://morphhoi.github.io}}
\authornote{\affil{\textdagger}Corresponding author.}

\begin{document}

\input{sections/teaser}
\input{sections/0_abstract}
\input{sections/1_intro}
\input{sections/2_related}
\input{sections/3_method}
\input{sections/4_experiments}

\input{sections/5_conclusion}

\bibliographystyle{plainnat}
\bibliography{main}

\input{sections/X_suppl}

\end{document}

%% file: sections/teaser.tex
\twocolumn[{%
\maketitle
\begin{center}
  \includegraphics[width=\textwidth]{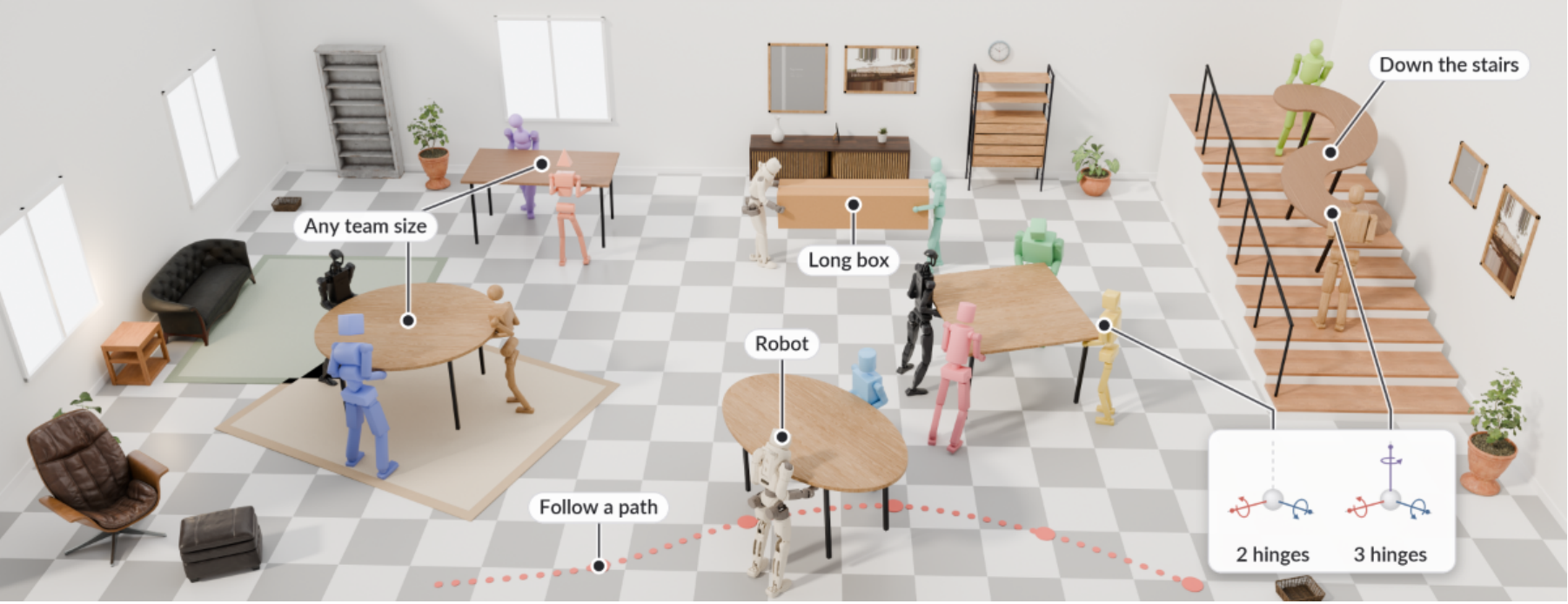}
  \vspace{-1.2em}
  \captionof{figure}{\textbf{One decentralised policy across embodiments, teams and skills.}
  \underline{Embodiments:} participants with diverse body proportions, skeletal structures, and joint configurations, including robot embodiments, are controlled by a single shared policy.
  \underline{Teams:} agents of any body form teams of any size around diverse tables.
  \underline{Skills:} skills added to the frozen cooperative policy carry an elliptical table along a path of waypoints, a box, and an S-shaped table down a flight of stairs. All agents are controlled by a single shared, decentralised policy.}
  \label{fig:teaser}
\end{center}
\vspace{0.8em}
}]

%% file: sections/0_abstract.tex
\begin{abstract}
Physics-based human-object interaction (HOI) has enabled multiple characters to cooperatively transport objects, but existing cooperative controllers largely assume homogeneous embodiments and predefined interaction tasks. We ask whether a single physics-based policy can enable cooperation across diverse embodiments, team compositions, and interaction skills. We present MorphHOI, a decentralised framework built on a key insight: effective heterogeneous cooperation requires a unified representation across embodiments while accounting for differences in their physical capabilities. To realise this insight, we introduce a canonical body interface that unifies observation and action spaces, together with an embodiment-conditioned transformer that adapts cooperative decisions to the physical capabilities of each agent and its teammates. To extend cooperation to new interaction skills, we reuse the learned cooperative policy through task-specific tokens and lightweight adaptation, preserving its original capabilities. Our framework achieves success rates of up to 96.7\% on seen embodiments, 95.6\% on unseen embodiments, and 91.4\% on novel payloads. Furthermore, the learned cooperation extends to diverse interaction skills with high success rates, including box carrying, transport on stairs, and waypoint following, while preserving previously acquired cooperative capabilities.
\end{abstract}

%% file: sections/1_intro.tex
\section{Introduction}
\label{sec:intro}

Cooperative human-object interaction (HOI) requires multiple agents to coordinate their actions towards shared goals while adapting to one another's physical capabilities. In practice, participants often differ in body proportions, joint configurations, and strengths, making such coordination particularly challenging. For instance, carrying a large object together requires partners to agree on where to grasp, when to lift, and how to move while sharing the load.
Physics-based HOI~\cite{peng2018deepmimic,peng2021amp} has progressed from individual manipulation skills~\cite{hassan2023synthesizing,wang2023physhoi,xiao2024unified,pan2025tokenhsi} to cooperative transport~\cite{gao2024coohoi,lionar2026teamhoi}. However, existing cooperative controllers largely assume homogeneous teams, limiting their ability to coordinate agents with diverse embodiments. Moreover, extending learned cooperative behaviours to new interaction skills often requires updating the shared policy, potentially compromising previously acquired capabilities~\cite{kirkpatrick2017overcoming}. These limitations raise a central question: \textbf{can a single policy enable physically plausible cooperation across heterogeneous embodiments and team compositions while remaining extensible to new interaction skills?}

Answering this question poses two closely related challenges. First, cooperative control must generalise across embodiments and team compositions. Characters with different skeletons have incompatible observation and action spaces~\cite{kurin2020my,gupta2022metamorph}, making it difficult to share a policy. More importantly, effective cooperation requires understanding both one's own capabilities and those of one's partners: characters of different heights and strengths must coordinate where to grasp, how high to lift, and how to move. This becomes harder as teams grow and diversify~\cite{bettini2023heterogeneous}. Second, cooperative skills should be unified and extensible across interaction tasks. Following routes, handling objects of different shapes, and traversing uneven terrain share fundamental cooperative behaviours but require different task-specific interactions. Training separate controllers limits the reuse of learned cooperation~\cite{xiao2024unified,pan2025tokenhsi}. A unified policy should therefore retain these shared capabilities while flexibly adapting to new interaction requirements.

We introduce \method, a framework for physics-based cooperative HOI across heterogeneous embodiments and extensible interaction skills. A canonical body interface unifies observations through the links of a shared skeleton and maps canonical joint targets to each character's own hinges, enabling one policy to control bodies with different skeletons and degrees of freedom (DoFs). Within this shared representation, an embodiment-conditioned transformer (ECT) accounts for differences in physical capabilities: each agent's own description conditions its network layers and action head, while teammate descriptions guide cooperative reasoning through corresponding tokens. Trained on mixed teams with body-adapted rewards and masked AMP~\cite{peng2021amp,lionar2026teamhoi} for motion imitation, the policy learns to coordinate agents with diverse capabilities. To extend this learned cooperation to new interaction skills, we condition the shared policy on task-specific tokens and introduce lightweight adaptation, enabling new behaviours while preserving the original cooperative policy.

With a single policy across diverse embodiments, \method achieves success rates of 96.7\%, 94.5\%, and 87.8\% with two, four, and eight agents, respectively. Across the same team sizes, it further generalises to unseen embodiments with success rates of 95.6\%, 93.9\%, and 80.0\%, and unseen objects with 90.0\%, 91.4\%, and 84.9\%, while extending to new skills with harmonic-mean success rates of 88.5\% to 92.2\% without compromising the original task. In summary, our contributions are:
\begin{itemize}

  \item \textbf{Unified control across heterogeneous embodiments.}
        We introduce a canonical body interface that unifies the observation and action spaces of characters with diverse body proportions and joint configurations, enabling a single decentralised policy to control heterogeneous teams.

  \item \textbf{Embodiment-conditioned cooperative reasoning.}
        We develop an embodiment-conditioned transformer that separately models inter-agent coordination and task interactions through structured attention, while conditioning cooperative decisions on the physical capabilities of each agent and its teammates.

  \item \textbf{Extensible cooperative interaction.}
We develop a progressive training framework that learns reusable cooperation across heterogeneous teams and extends it to new interaction skills by conditioning the shared policy on task-specific tokens with lightweight adaptation, while preserving previously learned cooperative behaviours.

\end{itemize}

%% file: sections/2_related.tex
\begin{figure*}[t]
  \centering
  \includegraphics[width=\textwidth]{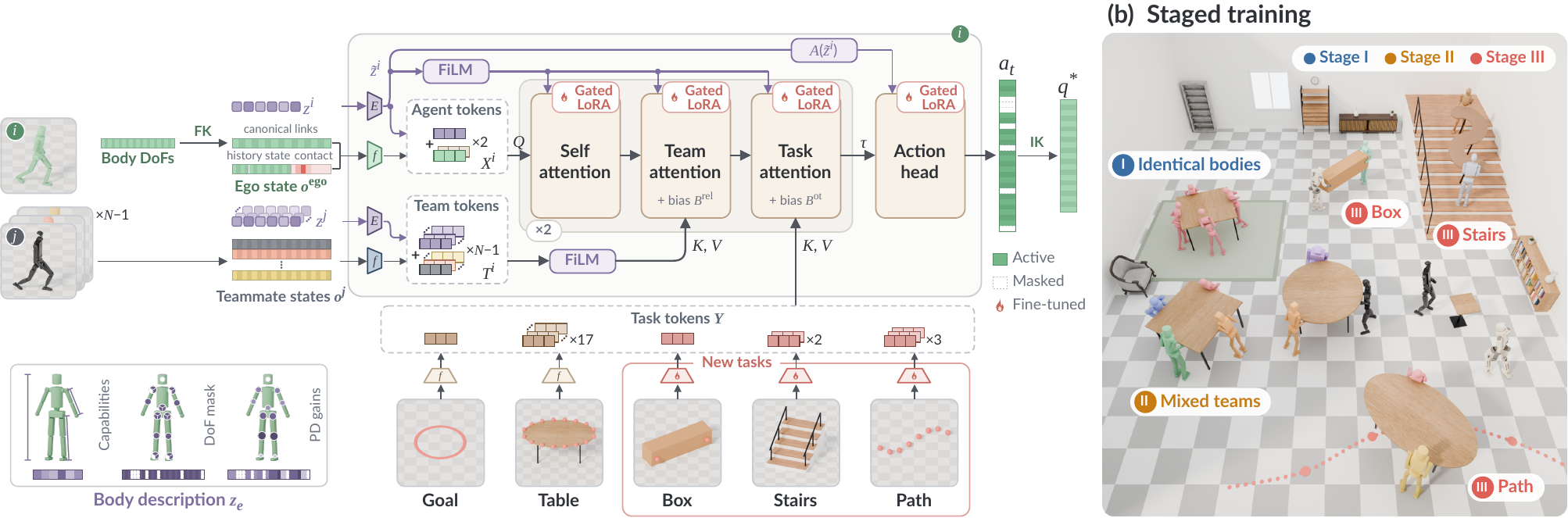}
  \vspace{-1.5em}
  \caption{\textbf{Overview of \method.} (a) Agent $i$ (green) observes its ego state $o^\mathrm{ego}$ via canonical links (FK) and teammate states $o^j$ relative to it. Tokenizers $f$ produce agent $X^i$ (queries $Q$), team $T^i$, and task $Y$ tokens (keys and values $K,V$). Encoder $E$ embeds body descriptions $z$ into corresponding tokens; the ego code $\tilde z^i$ modulates sublayers (FiLM) and generates $A(\tilde z^i)$. Team and task attention use biases $B^\mathrm{rel}$ and $B^\mathrm{ot}$. Readout $\tau$ produces canonical actions $a_t$, masked and mapped to hinge targets $q^\ast$ (IK). New skills add tokenizers and gated LoRA. (b) Stage I trains identical teams; Stage II learns heterogeneous cooperation; Stage III adds skills while preserving existing behaviours.}

  \label{fig:method}
\end{figure*}

\section{Related Work}
\label{sec:related}

\subsection{Cooperative Multi-Character Interaction}
\label{sec:rel_coop}

Single-character physics-based HOI, driven by motion imitation~\cite{peng2018deepmimic,luo2023perpetual,tessler2024maskedmimic} and adversarial motion priors~\cite{peng2021amp,peng2022ase,tessler2023calm} on captured motion~\cite{mahmood2019amass}, has progressed from individual skills~\cite{hassan2023synthesizing,wang2023physhoi,xu2025intermimic} to unified multi-task controllers~\cite{xiao2024unified,pan2025tokenhsi}. Recent task-conditioned controllers~\cite{pan2025tokenhsi,vainshtein2026task} enable diverse single-character skills, while extending such capabilities to multi-agent cooperative skills remains challenging due to the need for coordinated interaction among agents. Multi-character interaction is further constrained by limited interaction data~\cite{xu2024inter,zhang2024hoi,liu2025core4d,lee2026mochi}, motivating cooperative HOI methods to learn from single-person motion. CooHOI~\cite{gao2024coohoi} fine-tunes a single-agent carrying policy on multiple agents coordinating implicitly through the object. TeamHOI~\cite{lionar2026teamhoi} introduces teammate tokens and a masked motion prior, enabling a single decentralized policy to be applied to teams consisting of varying numbers of members. Recent work transfers solo skills to cooperation~\cite{yao2026synagent} or plans roles and object trajectories with a commander~\cite{huang2026scalable}. In robotics, humanoids carry objects with humans~\cite{agravante2019human,du2025learning} or each other~\cite{hawley2019control,lin2026cocarry}, and legged robots move payloads together~\cite{pandit2024learning,feng2025learning,cao2026paint}. Such systems pair varied participants but target specific platforms, while physics-based cooperative controllers largely assume homogeneous teams and predefined tasks. Coordinating diverse embodiments and capabilities under one policy, and extending this cooperation to new skills, remain open.

\subsection{Control across Embodiments}
\label{sec:rel_embodiment}

Morphology-agnostic control shares one policy across bodies by representing limbs as graphs processed by graph networks~\cite{wang2018nervenet,huang2020one} or transformers~\cite{kurin2020my,gupta2022metamorph,hong2021structure,sferrazza2024body}, or by conditioning shared networks on body descriptions through hardware encodings~\cite{chen2018hardware}, hypernetworks~\cite{ha2017hypernetworks,xiong2023universal}, and feature-wise modulation~\cite{perez2018film}. Universal policies extend these ideas to real robots, from legged locomotion~\cite{feng2023genloco,shafiee2024manyquadrupeds,bohlinger2024one,ai2025towards} to manipulation~\cite{patel2025get,doshi2024scaling}. For human-like bodies, simulated characters adapt to varying body shapes~\cite{won2019learning}, while humanoid controllers typically learn from retargeted human motion~\cite{araujo2026retargeting,yang2026omniretarget} for individual robots~\cite{he2024learning,he2024omnih2o,fu2024humanplus,cheng2024expressive,he2025hover,he2025asap,luo2026sonic}. Recent cross-humanoid controllers further share one policy through morphology randomisation~\cite{xue2026scalable} or shared motion tokens with robot-specific modules~\cite{zhang2026x}. While these approaches enable control across different embodiments, they do not explicitly address coordination among agents with different physical capabilities. Heterogeneous multi-agent learning~\cite{kuba2021trust,bettini2023heterogeneous,zhong2024heterogeneous} considers agent differences, but often relies on agent-specific policies. \method bridges these directions by enabling a single decentralised policy to coordinate heterogeneous agents based on their own physical capabilities and those of their teammates.

%% file: sections/3_method.tex
\section{Method}
\label{sec:method}

\subsection{Problem Formulation}
\label{sec:formulation}

A team of $N$ agents walks to an object, lifts, transports and puts it down at a target.
Agent $i$ controls embodiment $e_i \in \mathcal{E}$, characterised by a kinematic tree with $n_{e_i}$ actuated hinges, link masses and shapes, joint limits, and PD gains.
In this partially observable Markov game, each agent observes $o^i_t$ and outputs a body-dependent action $a^i_t \in \mathbb{R}^{n_{e_i}}$.
We seek a policy $\pi_\theta$ shared across all embodiments that maximises
$
  \mathbb{E}_{(e_1,\dots,e_N) \sim p(\mathcal{E}^N)}\; \mathbb{E}_{\pi_\theta}\Big[\tfrac{1}{N}\textstyle\sum_{i,t} \gamma^t r^i_t\Big],
$
where the outer expectation samples team compositions and $r^i_t$ is agent $i$'s reward (\cref{eq:reward}).
Since optimal actions depend on both ego and teammate embodiments, \method combines a shared canonical interface (\cref{sec:canonical}), team-conditioned cooperative control (\cref{sec:ect,sec:training}), and skill extension that preserves performance on the original task (\cref{sec:skills}).

\subsection{Unified Canonical Embodiment Interface}
\label{sec:canonical}

A single policy across heterogeneous embodiments requires unified observation and action spaces despite differences in skeletal structures and joint configurations. Meanwhile, shared representations must also preserve embodiment-specific physical capabilities to enable effective cooperation.

\subsubsection{Canonical Observation.}
Every embodiment has the 15 links of one canonical skeleton, including pelvis, torso, head, two arms and two legs of three links
each, but between two links it may have zero to three hinges, in any order and with offsets or tilted axes
(\cref{fig:canon_joint}). We thus adopt a unified link-level representation across different embodiments. With forward kinematics $\Phi_e$ of embodiment
$e$, the agent's canonical link state is
\begin{equation}
  s^i_t = \Phi_{e_i}(q^i_t, \dot q^i_t) = \big(h_t,\, \{x_l, R_l, \dot x_l, \omega_l\}_{l=1}^{15}\big),
  \label{eq:canon_obs}
\end{equation}
where $h_t$ is the root height and $x_l, R_l, \dot x_l, \omega_l$ are the position relative to the root, rotation,
linear and angular velocity of link $l$, all in the root's heading frame.

\subsubsection{Canonical Action.}
\begin{figure}[t]
  \centering
  \includegraphics[width=\columnwidth]{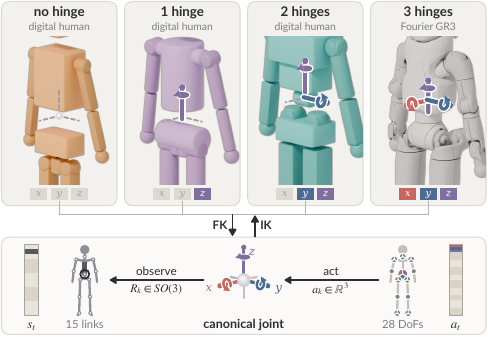}
  \vspace{-1.6em}
  \caption{\textbf{One canonical joint, many hinge chains.} Waist joints of four seen bodies with 0 to 3 hinges
(solid: hinges; dashed: absent canonical axes; below: DoF mask $m_e$).
Torso rotation $R_k$ is observed via FK; action $a_k \in \mathbb{R}^3$ is mapped via IK.}
  \label{fig:canon_joint}
\end{figure}

The policy outputs $a \in \mathbb{R}^{28}$, PD targets for the 12 actuated canonical joints, as illustrated in \cref{fig:canon_joint}. For a three-DoF joint $k$ with target rotation $R^*_k = \exp([a_k]_\times)$ and the body's hinge chain
$R_k(q) = \prod_n \exp([\hat\omega_n]_\times q_n)$ over its hinges $n$ with unit axes $\hat\omega_n$,
the hinge targets $q^*$ follow from damped Gauss-Newton steps
\begin{equation}
  q \leftarrow \Pi\Big(q + (J^\top J + \lambda I)^{-1} J^\top \log\big(R^*_k\, R_k(q)^\top\big)^\vee\Big),
  \label{eq:ik}
\end{equation}
warm-started at the previous targets, where $J$ is the spatial Jacobian of the chain and $\Pi$ clips to the
intersection of the hinge and canonical limits. Canonical DoFs absent from an embodiment are masked by $m_e \in \{0,1\}^{28}$:
their actions are set to zero and excluded from the log-likelihood $\sum_d m_{e,d} \log \mathcal{N}(a_d; \mu_d, \sigma_d^2)$.
The same canonical action thus produces the same joint rotation on every body.

\subsubsection{Embodiment Description.}
To enable generalisation across embodiments, the policy conditions on physical characteristics and capabilities,
\begin{equation}
  z_e = \big[\, c_e,\; m_e,\; \log(k_e / k_0) \,\big],
  \label{eq:ze}
\end{equation}
where $c_e$ holds capability features measured against the reference human~\cite{peng2021amp}, including log stature, arm and leg length ratios,
mass and joint strength relative to a human of that stature; $m_e \in \{0,1\}^{28}$ marks which of the
reference human's 28 DoFs embodiment $e$ can actuate; and $k_e / k_0$ is the per-DoF ratio of the body's PD
stiffness to the reference human's. Similar bodies thus share what they learn, and an unseen embodiment is described in the
same terms.

\subsubsection{Motion Prior for All Embodiments.}
The discriminator features $\hat s_t$ describe every body by its canonical joint rotations
$\hat q_k = \log(R_{p(k)}^\top R_{c(k)})^\vee$ between parent and child links, with link states scaled to the reference
stature. Following prior work~\cite{peng2021amp,lionar2026teamhoi}, a discriminator $D_\phi$ scores transitions
of these features against reference motion, yielding the style reward
$r^\mathrm{style}_t = -\log(1 - D_\phi(\hat s_t, \hat s_{t+1}))$.
Every body is thus scored against the same reference motion; as in TeamHOI~\cite{lionar2026teamhoi},
the arms are masked near the object.
A human gait imposed on a body with other proportions and gains destabilises it, so the
style reward is weighted per body,
\begin{equation}
  r^i_t = w_G\, r^{\mathrm{task},i}_t + w_S\, w_{e_i}\, r^{\mathrm{style},i}_t ,
  \label{eq:reward}
\end{equation}
where $w_G$ and $w_S$ balance task and style. We optimise the shared policy with PPO~\cite{schulman2017proximal}.

\subsection{Embodiment-Conditioned Transformer}
\label{sec:ect}

Although agents share canonical states, their actions depend on their own and teammates' embodiments. ECT (\cref{fig:method}a) captures these dependencies through embodiment-aware tokens, structured attention, and body-specific modulation. We write $z^i = z_{e_i}$, $\tilde z^i = \tilde z_{e_i}$ and omit time indices for brevity.

\subsubsection{Embodiment-Aware Tokens.}
In agent $i$'s heading frame, the ego observation
$o^\mathrm{ego} = (s^i, \bar s^i, \hat P_i)$ contains the canonical link state of \cref{eq:canon_obs}, a four-frame history $\bar s^i$ of root and hand motion, and the agent's contact assignment $\hat P_i$. Here, $\hat P_i$ is row $i$ of the team contact assignment $\hat P$.
Each teammate $j$ contributes $o^j = (\bar s^j, \hat P_j)$, containing its motion history and contact assignment relative to agent $i$.
Task observations comprise the object state $o^\mathrm{obj}$, $M$ rim contact points $p_m$, and the goal $o^\mathrm{goal}$.
We use shared tokenizers $f_\ast$ and two embodiment encoders: $E(z)$ embeds physical characteristics into tokens, while $\tilde z = \tilde E(z)$ provides conditioning features. Both encoders are shared across bodies and shown as $E$ in \cref{fig:method}.
We define feature-wise modulation as
$\mathrm{FiLM}(x;\tilde z)=(1+\zeta)\odot x+\xi$~\cite{perez2018film},
where $(\zeta,\xi)$ are generated by zero-initialised linear projections of $\tilde z$, with separate parameters for each application.
The three token groups are
\begin{equation}
\begin{aligned}
  X^i &= \big[\,\tau,\, f_\mathrm{ego}(o^\mathrm{ego}) + E(z^i)\,\big], \\
  T^i &= \big[\,\mathrm{FiLM}\big(f_\mathrm{mate}(o^j) + E(z^j);\, \tilde z^j\big)\,\big]_{j \neq i}, \\
  Y &= \big[\,f_\mathrm{obj}(o^\mathrm{obj}),\,
  \{f_\mathrm{con}(p_m)\}_{m=1}^{M},\,
  f_\mathrm{goal}(o^\mathrm{goal})\,\big].
\end{aligned}
\label{eq:tokens}
\end{equation}
Here, $X^i$, $T^i$, and $Y$ are the agent, team, and task tokens, respectively, and $\tau$ is a learned readout token. Each teammate token combines behaviour, assigned contact, and capabilities, with FiLM adapting its features to the corresponding body. This lets the policy distinguish what teammates are doing from what their embodiments allow them to do.

\begin{table*}[t]
  \centering
  \setlength{\tabcolsep}{2.4pt}
  \resizebox{\textwidth}{!}{\input{tables/table1_main}}
  \vspace{-0.6em}
  \caption{\textbf{Teams of different bodies carrying a table}; all methods train on two to four agents with the exception of TeamHOI. Avg: mean over team
  sizes. Where the SR is 0, other metrics are not reported (-). $^*$Retrained on our bodies (\cref{sec:setup}).}
  \label{tab:main}
\end{table*}

\begin{figure*}[t]
  \centering
  \includegraphics[width=\textwidth]{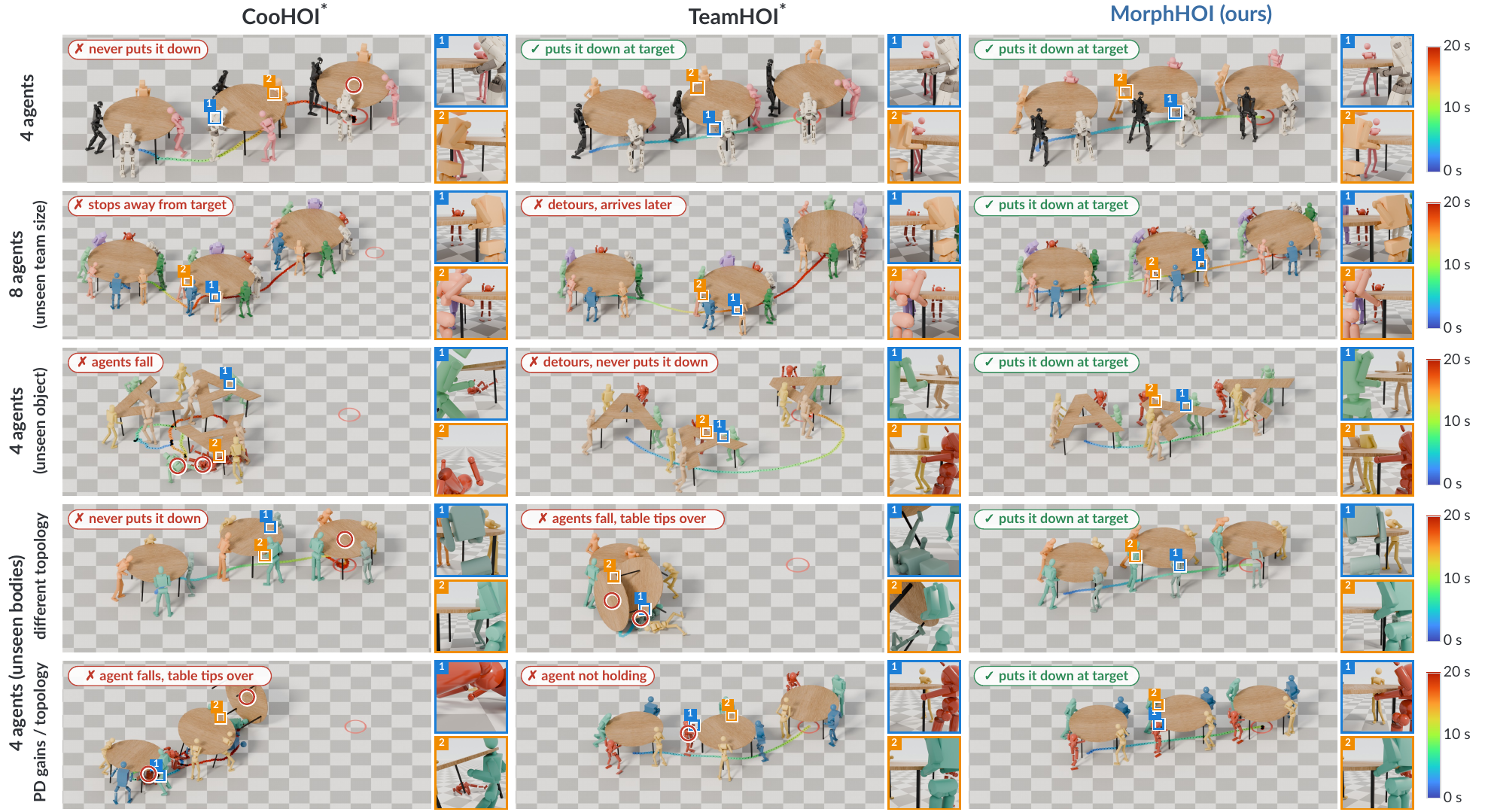}
  \vspace{-1.5em}
  \caption{\textbf{Qualitative comparison} under identical teams, spawns and targets: four seen embodiments; eight seen embodiments; an unseen A-shaped table; four unseen embodiments (all with unseen topologies); and four unseen embodiments (two with unseen PD gains and two with unseen topologies). Each tile shows three moments,
  the object path coloured by time and the target as a red ring; close-ups (1, 2) show hand-object contact.}
  \label{fig:compare}
\end{figure*}

\subsubsection{Structured Attention.}
Unlike TeamHOI~\cite{lionar2026teamhoi}, which attends to teammates within a shared attention stack, ECT separates self-interaction, teammate coordination, and task reasoning to model their distinct dependencies.
Let $\mathcal{A}(Q,K;B)$ denote multi-head attention with queries $Q$, keys and values from $K$, and optional logit bias $B$.
Each of the $L$ transformer blocks sequentially applies
\begin{equation}
\begin{aligned}
  X^i &\leftarrow \mathcal{A}_\mathrm{self}\big(X^i,\, X^i\big), \\
  X^i &\leftarrow \mathcal{A}_\mathrm{team}\big(X^i,\, T^i;\, B^\mathrm{rel}\big), \\
  X^i &\leftarrow \mathcal{A}_\mathrm{task}\big(X^i,\, Y;\, B^\mathrm{ot}\big).
\end{aligned}
\label{eq:blocks}
\end{equation}
Each operation has independent attention parameters and a feed-forward sublayer, both embodiment-conditioned as described below.
By retaining $X^i$ as queries, ECT accommodates varying team sizes and additional skill tokens (\cref{sec:skills}) while separately reasoning about teammates and task constraints.
Spatial relationships and grasping responsibilities enter through two attention biases:
\begin{equation}
  B^\mathrm{rel}_{h,j} = g_h(r^{ij}), \qquad
  B^\mathrm{ot}_{h,m} = \beta_h \log \hat P_{i,m},
  \label{eq:bias}
\end{equation}
where $h$ indexes attention heads, $g_h$ is an MLP evaluating the inter-agent relation $r^{ij}$ (distance, closing speed, and contact-assignment overlap), and $\beta_h$ is learned.
$B^\mathrm{rel}$ biases teammate attention, whereas $B^\mathrm{ot}$ biases the $M$ contact-point tokens, with zero bias for other task tokens.
The assignment $\hat P \in \mathbb{R}^{N \times M}$ is obtained by row-normalising an entropic optimal transport solution~\cite{cuturi2013sinkhorn} between $N$ agents and $M$ contacts, using embodiment-dependent scores and capacities (\cref{sec:app_ot}).
It allocates grasping responsibilities according to physical capabilities, guiding coordinated interaction.

\subsubsection{Embodiment-Conditioned Modulation.}
While teammate descriptions guide cooperation, the ego embodiment determines how shared features become actions.
We condition each attention and feed-forward sublayer $F$ on $\tilde z^i$ through FiLM and a residual gate:
\begin{equation}
  x \leftarrow \mathrm{FiLM}\big(
  \mathrm{LN}\big(x + (1+\eta)\odot F(x)\big);
  \,\tilde z^i\big),
  \label{eq:film}
\end{equation}
where $\mathrm{LN}$ is layer normalisation and $\eta$ is generated by a separate zero-initialised linear projection of $\tilde z^i$ for each sublayer.
The gate adjusts the residual contribution, while FiLM modulates its output, adapting shared computation to the ego body's capabilities.
After $L$ blocks, the readout feature $v$ gives the Gaussian policy mean $\mu$ over canonical actions:
\begin{equation}
  \mu = m_{e_i} \odot
  \big((W_0 + A(\tilde z^i) B)\, v
  + b_0 + c(\tilde z^i)\big),
  \label{eq:hyper}
\end{equation}
where $W_0,b_0$ are shared parameters, $B$ is a shared rank-$r$ projection, and $A(\tilde z^i),c(\tilde z^i)$ are zero-initialised linear maps producing embodiment-specific corrections~\cite{ha2017hypernetworks}.
Thus, the low-rank head adapts action generation to each body's motor capabilities, while $m_{e_i}$ masks absent canonical DoFs.

\subsection{Learning Heterogeneous Cooperation}
\label{sec:training}

\subsubsection{Embodiment-Adapted Rewards.}
To accommodate heterogeneous capabilities, we adapt the training rewards. Building on the task reward of TeamHOI~\cite{lionar2026teamhoi}, we introduce three adaptations: normalised lifting progress provides gradual feedback across different lifting capabilities; signed transport progress rewards motion towards the target while all agents maintain contact; and a put-down term $\exp(-k_z \Delta z)$ encourages lowering the object to its rest height after arrival (see \cref{sec:app_rewards}). We further adapt the hand-rim contact threshold to each embodiment. For hands hanging $\Delta h_e$ below the rim when standing, we define
\begin{equation}
  \delta_e = \delta_0 + \min(\Delta h_e, \Delta h_\mathrm{max}),
  \label{eq:gate}
\end{equation}
where $\delta_0$ is the contact threshold for the reference human and $\Delta h_\mathrm{max}$ caps the adjustment. This accounts for differences in body proportions, preventing shorter agents from being penalised for their limited reach.

\subsubsection{From Homogeneous to Heterogeneous Teams.}
A team in which any body is still unstable rarely completes a carry, since the fall of one agent ends the episode.
Stage I therefore trains on teams of one body, where every agent learns to walk, lift, carry and put down with
partners like itself; \method is trained from scratch with PPO~\cite{schulman2017proximal}. Stage II continues on free combinations of
the training bodies in a share $\rho_\mathrm{mix}$ of the environments, so that each agent meets teammates whose size,
gait and strength differ from its own. Only in such teams do the partners' descriptions carry information that the
agent's own does not, so stage II is where the policy learns to use them.

\subsection{Task-Conditioned Skill Extension}
\label{sec:skills}

A new skill should reuse the cooperative, embodiment-aware behaviour of stage II, yet must not change it for agents
that do not use the skill. A skill $s$, such as following a path, carrying a long box or traversing stairs, brings new
observations (the next waypoints, the box and its grasp points, the ground ahead). Stage III tokenises them with a new tokenizer $f_s$,
appends them to the task tokens $Y$, and gives every linear map $W$ of every sublayer and of the action head a
low-rank update~\cite{hu2021lora} gated per agent,
\begin{equation}
  W x \;\rightarrow\; W x + g_i\, B_s A_s x,
  \label{eq:lora}
\end{equation}
where $A_s$ and $B_s$ are rank-$r_s$ factors ($B_s$ zero-initialised),
and $g_i$ activates the update when skill $s$ is assigned to agent $i$, whose task observations are encoded by the corresponding tokenizer.
All other actor weights, FiLM and $A(\tilde z)$ remain
frozen, as in progressive networks~\cite{rusu2016progressive}. For an agent with $g_i = 0$ the updates are bypassed and the
skill tokens are masked out; with the observation normaliser of stage II, it computes exactly the stage II policy. Therefore, the performance on the base task does not degrade.

%% file: tables/table1_main.tex
\begin{tabular}{@{}ll cccc c cccc c cccc c cccc@{}}
\toprule
 & & \multicolumn{4}{c}{2 agents} && \multicolumn{4}{c}{4 agents} && \multicolumn{4}{c}{8 agents (unseen team size)} && \multicolumn{4}{c}{Avg} \\
\cmidrule{3-6}\cmidrule{8-11}\cmidrule{13-16}\cmidrule{18-21}
Setting & Method & SR$\uparrow$ & DE$\downarrow$ & CR$\uparrow$ & SPARC$\uparrow$ && SR$\uparrow$ & DE$\downarrow$ & CR$\uparrow$ & SPARC$\uparrow$ && SR$\uparrow$ & DE$\downarrow$ & CR$\uparrow$ & SPARC$\uparrow$ && SR$\uparrow$ & DE$\downarrow$ & CR$\uparrow$ & SPARC$\uparrow$ \\
\midrule
\multirow{5}{*}{Seen bodies} & TeamHOI~\cite{lionar2026teamhoi} & 0.0 & -- & -- & -- && 0.0 & -- & -- & -- && 0.0 & -- & -- & -- && 0.0 & -- & -- & -- \\
 & CooHOI$^*$~\cite{gao2024coohoi} & 89.9 & 0.49 & 83.6 & $-$2.23 && \underline{90.8} & \underline{0.48} & 76.7 & $-$2.06 && 54.3 & 3.07 & 17.5 & $-$2.16 && 78.3 & 1.35 & 59.3 & $-$2.15 \\
 & TeamHOI$^*$~\cite{lionar2026teamhoi} & \underline{95.3} & \underline{0.34} & \underline{88.1} & \underline{$-$1.99} && 88.0 & 0.84 & \textbf{90.2} & \underline{$-$1.96} && \underline{81.7} & 1.01 & \underline{46.0} & \underline{$-$2.12} && \underline{88.3} & 0.73 & \underline{74.8} & \underline{$-$2.02} \\
 & \method, stage I & 90.3 & 0.49 & 82.9 & $-$2.04 && 90.2 & 0.55 & 73.6 & $-$1.99 && 78.8 & \underline{0.89} & 42.1 & $-$2.31 && 86.4 & \underline{0.64} & 66.2 & $-$2.11 \\
 & \method, stage II & \textbf{96.7} & \textbf{0.24} & \textbf{93.8} & \textbf{$-$1.85} && \textbf{94.5} & \textbf{0.31} & \underline{86.3} & \textbf{$-$1.92} && \textbf{87.8} & \textbf{0.60} & \textbf{58.7} & \textbf{$-$1.98} && \textbf{93.0} & \textbf{0.38} & \textbf{79.6} & \textbf{$-$1.92} \\
\midrule
\multirow{5}{*}{Unseen embodiments} & TeamHOI~\cite{lionar2026teamhoi} & 0.0 & -- & -- & -- && 0.0 & -- & -- & -- && 0.0 & -- & -- & -- && 0.0 & -- & -- & -- \\
 & CooHOI$^*$~\cite{gao2024coohoi} & 88.4 & 0.58 & 69.5 & $-$2.30 && \underline{88.1} & 0.65 & 62.9 & $-$2.09 && 46.7 & 3.58 & 12.7 & $-$2.56 && 74.4 & 1.60 & 48.4 & $-$2.32 \\
 & TeamHOI$^*$~\cite{lionar2026teamhoi} & \underline{93.5} & \textbf{0.34} & \underline{73.8} & \underline{$-$2.00} && 87.5 & 0.96 & \textbf{79.7} & \underline{$-$1.97} && \underline{73.8} & 1.32 & \underline{32.2} & \underline{$-$2.14} && \underline{84.9} & 0.87 & \underline{61.9} & \underline{$-$2.04} \\
 & \method, stage I & 87.3 & 0.56 & 67.9 & $-$2.15 && 87.8 & \underline{0.59} & 59.3 & $-$2.04 && 73.4 & \underline{0.97} & 30.9 & $-$2.36 && 82.8 & \underline{0.71} & 52.7 & $-$2.18 \\
 & \method, stage II & \textbf{95.6} & \underline{0.40} & \textbf{81.8} & \textbf{$-$1.90} && \textbf{93.9} & \textbf{0.33} & \underline{76.3} & \textbf{$-$1.93} && \textbf{80.0} & \textbf{0.96} & \textbf{44.0} & \textbf{$-$2.09} && \textbf{89.8} & \textbf{0.56} & \textbf{67.4} & \textbf{$-$1.97} \\
\midrule
\multirow{5}{*}{Unseen object} & TeamHOI~\cite{lionar2026teamhoi} & 0.0 & -- & -- & -- && 0.0 & -- & -- & -- && 0.0 & -- & -- & -- && 0.0 & -- & -- & -- \\
 & CooHOI$^*$~\cite{gao2024coohoi} & 78.8 & 1.05 & 75.0 & $-$2.28 && \underline{86.1} & \underline{0.78} & 71.1 & $-$2.15 && 46.5 & 3.30 & 8.8 & $-$2.53 && 70.5 & 1.71 & 51.6 & $-$2.32 \\
 & TeamHOI$^*$~\cite{lionar2026teamhoi} & \underline{86.6} & \underline{0.76} & \underline{81.9} & \textbf{$-$2.01} && 85.2 & 1.16 & \underline{75.2} & $-$2.02 && \underline{78.8} & \underline{1.01} & \underline{39.0} & \underline{$-$2.17} && \underline{83.5} & \underline{0.98} & \underline{65.4} & \underline{$-$2.07} \\
 & \method, stage I & 79.3 & 1.10 & 72.7 & \underline{$-$2.11} && 85.3 & 0.85 & 69.6 & \underline{$-$2.01} && 76.2 & 1.06 & 38.8 & $-$2.42 && 80.3 & 1.00 & 60.4 & $-$2.18 \\
 & \method, stage II & \textbf{90.0} & \textbf{0.62} & \textbf{82.9} & \textbf{$-$2.01} && \textbf{91.4} & \textbf{0.51} & \textbf{79.1} & \textbf{$-$1.93} && \textbf{84.9} & \textbf{0.61} & \textbf{54.9} & \textbf{$-$2.05} && \textbf{88.8} & \textbf{0.58} & \textbf{72.3} & \textbf{$-$2.00} \\
\bottomrule
\end{tabular}

%% file: sections/4_experiments.tex
\begin{figure}[!t]
  \centering
  \includegraphics[width=\columnwidth]{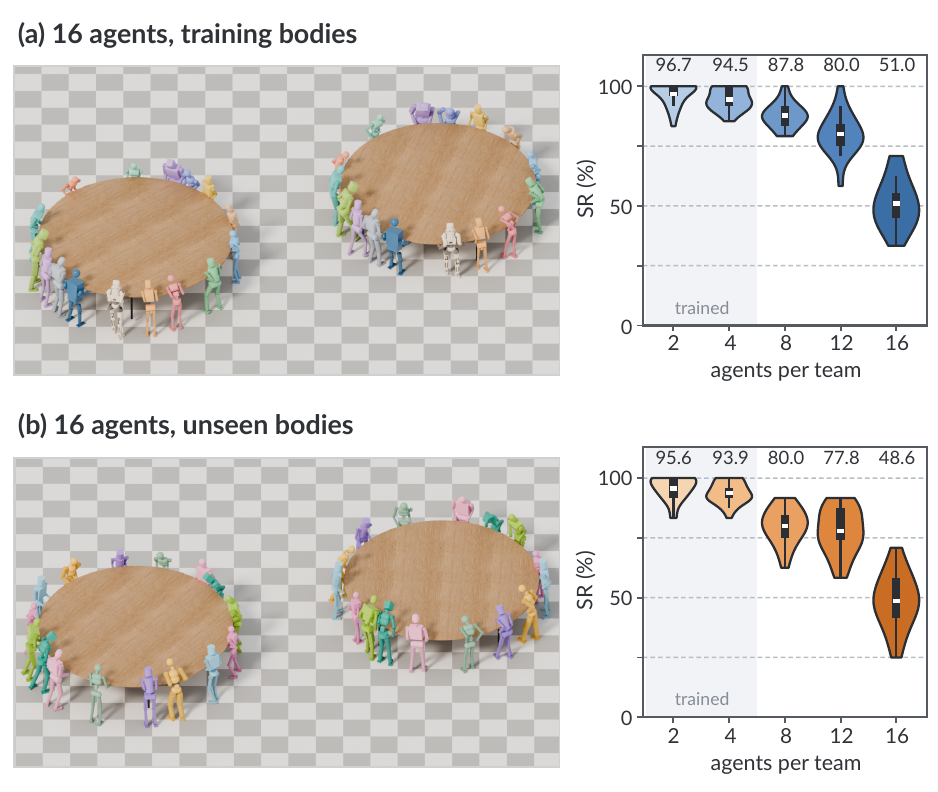}
  \vspace{-2.2em}
  \caption{\textbf{Larger teams}, (a) seen bodies and (b) unseen embodiments. Left: sixteen agents mid-carry
  and at the target. Right: SR against team size (violin: spread over team compositions; number: SR over all
  episodes; shaded: training sizes).}
  \label{fig:teamsize}
\end{figure}

\begin{figure*}[t]
  \centering
  \includegraphics[width=\textwidth]{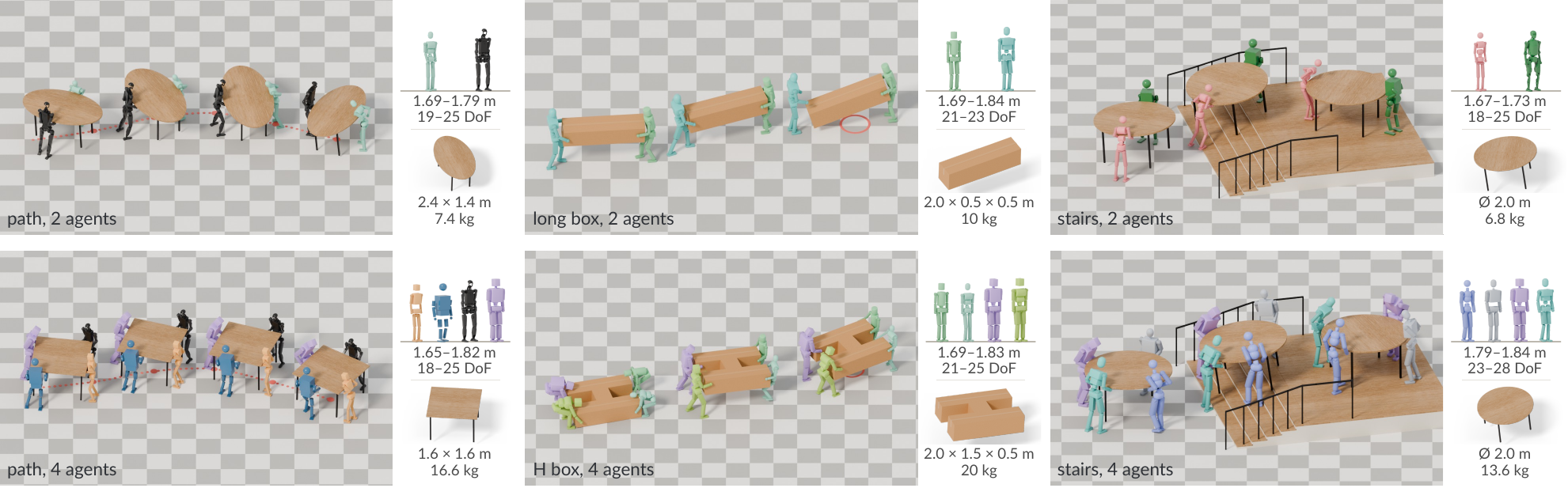}
  \vspace{-1.5em}
  \caption{\textbf{One policy, new skills.} Paths of waypoints (red), a long box and an H-shaped box, and a table carried down stairs, all added in
  stage III to the frozen policy, with teams of two (top) and four (bottom), the team sizes evaluated in
  \cref{tab:abl_skill}. Every pose is a rollout. Right of each tile: the team to scale, its range of stature
  and DoFs, and the object with its size and mass.}
  \label{fig:showcase}
\end{figure*}

\section{Experiments}
\label{sec:experiments}

\subsection{Experimental Setup}
\label{sec:setup}
All bodies are simulated humanoids, procedurally generated digital humans mixed with robot morphologies, whose skeletons differ widely (\cref{sec:app_bodies}). A team walks to a table, lifts it, carries it and puts it down at a target; training uses teams of two to four. Every evaluated team consists of different bodies, in three settings: \underline{seen bodies}, fixed teams of two, four and eight training bodies; \underline{unseen embodiments}, where embodiments not encountered during training with distinct morphologies form the whole team (\cref{sec:app_bodies}); and \underline{unseen objects}, a diverse set of previously unseen object geometries. All methods run same team setting for evaluation (\cref{sec:app_eval}). Ablations report the mean over the two-, four- and eight-agent team sets.
TeamHOI~\cite{lionar2026teamhoi} is the released policy evaluated zero-shot. To accommodate bodies with different joint spaces, we adapt both baselines, including the CooHOI$^*$ and TeamHOI$^*$, through our canonical body interface (\cref{sec:canonical}), providing a shared observation and action space while preserving their original architectures. TeamHOI$^*$ retains TeamHOI's transformer, rewards, and training, while CooHOI$^*$ follows the multi-agent adaptation of CooHOI~\cite{gao2024coohoi} provided by TeamHOI~\cite{lionar2026teamhoi}. Both are retrained on the same heterogeneous bodies and training episodes as our method. Following TeamHOI~\cite{lionar2026teamhoi}, the \underline{success rate} (SR, \%, primary) measures episodes where the table is placed at the target near its rest height without agent falls; \underline{distance error} (DE, m) measures its final distance to the target; \underline{cooperation ratio} (CR, \%) is the fraction of transport during which all agents maintain contact; and \underline{SPARC}, the spectral arc length~\cite{balasubramanian2015analysis} of the table's speed, measures smoothness with less sensitivity to contact noise than jerk.

\subsection{Generalisation across Embodiments}
\label{sec:exp_who}

The released TeamHOI policy, trained on a single digital human, fails to transfer across embodiments (SR 0.0\%, \cref{tab:main}). Retrained through our canonical interface, both baselines learn to cooperate but exhibit distinct limitations: CooHOI$^*$ struggles with implicit coordination through the object, while TeamHOI$^*$ exhibits detours, incomplete placement, and lost contact (\cref{fig:compare}). On seen bodies, averaged over team sizes, \method achieves 93.0\% SR against 88.3\% for TeamHOI$^*$ and 78.3\% for CooHOI$^*$, with lower distance error and smoother carries. With entirely unseen embodiments, \method retains 95.6\%, 93.9\%, and 80.0\% SR across two-, four-, and eight-agent teams, averaging 89.8\% against 84.9\% and 74.4\% for the baselines. Together with the ablations (\cref{tab:abl_cond}), these results highlight the importance of conditioning cooperation on both the agent's and its teammates' physical capabilities, beyond shared observations and actions.

\subsection{Generalisation across Teams and Objects}
\label{sec:exp_whom}

\subsubsection{Generalisation to Larger Teams.}
Although trained only with two to four agents, \method generalises to substantially larger teams. With eight agents, it achieves 87.8\% SR and 58.7\% CR, outperforming TeamHOI$^*$ (81.7\%, 46.0\%) and CooHOI$^*$ (54.3\%, 17.5\%). The particularly low CR of CooHOI$^*$ suggests that coordination through the shared object alone becomes insufficient as more agents must maintain contact and synchronise their movements. \cref{fig:teamsize} further shows that \method maintains 80.0\% SR with twelve agents, but drops to 51.0\% with sixteen, highlighting the increasing difficulty of coordinating larger teams. Nevertheless, unseen embodiments achieve comparable rates of 77.8\% and 48.6\%, respectively. The consistent performance across seen and unseen bodies suggests that the policy generalises across team sizes and diverse embodiment compositions simultaneously.

\subsubsection{Generalisation to Unseen Objects.}
On unseen objects, \method achieves success rates of 90.0\%, 91.4\%, and 84.9\% with two, four, and eight agents, averaging 88.8\% against 83.5\% for TeamHOI$^*$ and 70.5\% for CooHOI$^*$. Although TeamHOI$^*$ learns object-adaptive formations and is retrained on heterogeneous bodies, its shared transformer must infer embodiment-dependent constraints from agent states alone. New object geometries alter contact locations, imposing different reaching and support demands on agents with varying physical capabilities. By explicitly incorporating these capabilities into teammate representations and contact assignment, \method better adapts cooperation to both object geometry and team composition. In \cref{fig:compare}, CooHOI$^*$ loses stability and TeamHOI$^*$ fails to put down the object, whereas \method maintains coordinated contact and completes the transport.

\begin{table}[t]
  \centering
  \footnotesize
  \setlength{\tabcolsep}{1.6pt}
  \resizebox{\columnwidth}{!}{\input{tables/table2_abl_skill}}
  \vspace{-0.5em}
  \caption{\textbf{Adding a skill} (same tokens, environments and epochs per row; stairs: down).
    SR$_\mathrm{H}$, CR$_\mathrm{H}$: harmonic means over the skill and the table task without the skill input,
    each averaged over teams of two and four agents.}
  \label{tab:abl_skill}
\end{table}

\begin{table}[t]
  \centering
  \footnotesize
  \setlength{\tabcolsep}{1.6pt}
  \resizebox{\columnwidth}{!}{\input{tables/table3_abl_train}}
  \vspace{-0.5em}
  \caption{\textbf{Training for mixed teams}: stage II rerun from the same stage-I policy for the same
    epochs, one change per row.}
  \label{tab:abl_train}
\end{table}

\begin{table}[t]
  \centering
  \footnotesize
  \setlength{\tabcolsep}{1.6pt}
  \resizebox{\columnwidth}{!}{\input{tables/table4_abl_cond}}
  \vspace{-0.5em}
  \caption{\textbf{Body conditioning}: what the policy is told about its own and its teammates' bodies,
    and where $z_e$ enters. A body ID has no entry for a held-out embodiment (-).}
  \label{tab:abl_cond}
\end{table}

\subsection{Extending Cooperative Skills}
\label{sec:exp_what}

Stage III adds three skills to the frozen stage-II policy: box carrying, following five waypoints, and carrying a table down four steps of 2--12\,cm. Each skill is evaluated with two- and four-agent teams, averaged in \cref{tab:abl_skill}, with rollouts in \cref{fig:showcase}.
We compare four adaptation strategies: \underline{full fine-tune} updates all weights; \underline{LoRA w/o gate} sets $g_i=1$ for all agents in \cref{eq:lora}; \underline{tokenizer only} trains the new tokenizer with the actor frozen; and \underline{gated LoRA} is \method.
Performance is measured by the harmonic mean across the new skill and original table task, penalising failure on either~\cite{xian2017zero}.
Gated LoRA achieves the highest SR$_\mathrm{H}$ on all skills (88.6\%, 92.2\%, 88.5\%). Ungated updates interfere with the original task, whereas tokenizer-only adaptation preserves it but fails on box carrying, which requires supporting the object with both arms rather than grasping its rim. Even on the more similar path and stair tasks, alternatives trail by 8.3--12.1 points.
Per-agent gating resolves this trade-off by enabling skill-specific updates without altering the base policy for agents without the skill.
The same policy also supports teammates joining or leaving mid-carry and robot teams (\cref{sec:app_comego,sec:app_robots}).

\subsection{Ablation Studies}
\label{sec:exp_ablation}

\subsubsection{Embodiment-Aware Control.}
Each ablation retrains with the full recipe and averages over the teams in \cref{tab:main}; additional ablations are in \cref{sec:app_ect,sec:app_iface,sec:app_scale}. \Cref{tab:abl_cond} examines what embodiment information matters and how it enters the policy. Removing body descriptions while retaining the DoF mask reduces SR from 93.0\% to 80.5\% and increases DE from 0.38 to 1.02\,m, showing that a shared action space alone cannot ensure effective heterogeneous cooperation. Replacing physical descriptions with body IDs improves SR to 84.0\% but remains well below \method and cannot represent unseen bodies, highlighting the importance of transferable capability information over memorised identities. Removing the hyper-head (\cref{eq:hyper}) or FiLM reduces SR to 87.5\% and 90.2\%, respectively, suggesting that embodiment awareness benefits both action generation and intermediate feature adaptation.

\subsubsection{Learning Heterogeneous Cooperation.}
Stage II improves over stage I across all settings in \cref{tab:main}, highlighting the importance of mixed-team training. \Cref{tab:abl_train} isolates two complementary adaptations. Without the adaptive gate (\cref{eq:gate}), CR drops from 79.6\% to 73.0\%, as uniform contact thresholds disadvantage shorter-reach bodies. Without style reweighting, CR remains high (77.8\%), but SR falls to 85.7\% and distance error doubles. These results suggest that contact adaptation sustains cooperation, while embodiment-aware style weighting preserves transport precision despite differences in body dynamics.

\subsubsection{Teammate-Aware Cooperation.}
Removing teammate $z_e$ retains the agent's own description and its teammates' observed states, but removes explicit information about their physical capabilities. SR drops from 93.0\% to 89.4\%, while distance error more than doubles and carries become less smooth. This suggests that observing teammates' movements alone is insufficient to anticipate their reaching and load-sharing capabilities, leading to less precise coordination despite continued cooperation. Moreover, replacing each agent's own description with a teammate's reduces SR to 71.2\% (\cref{sec:app_zswap}), showing that embodiment information must be correctly associated with each participant to guide effective control.

%% file: tables/table2_abl_skill.tex
\begin{tabular}{@{}l cc c cc c cc@{}}
\toprule
 & \multicolumn{2}{c}{Box} && \multicolumn{2}{c}{Path} && \multicolumn{2}{c}{Stairs} \\
\cmidrule{2-3}\cmidrule{5-6}\cmidrule{8-9}
Method & SR$_\mathrm{H}{\uparrow}$ & CR$_\mathrm{H}{\uparrow}$ & & SR$_\mathrm{H}{\uparrow}$ & CR$_\mathrm{H}{\uparrow}$ & & SR$_\mathrm{H}{\uparrow}$ & CR$_\mathrm{H}{\uparrow}$ \\
\midrule
Full fine-tune & 0.0 & -- && 83.9 & 76.7 && 78.5 & 73.3 \\
LoRA w/o gate & 0.0 & -- && 83.7 & 74.3 && 77.2 & 68.6 \\
Tokenizer only & 0.0 & -- && 83.4 & 76.1 && 76.4 & 72.0 \\
Gated LoRA (\method) & \textbf{88.6} & \textbf{78.2} && \textbf{92.2} & \textbf{79.1} && \textbf{88.5} & \textbf{78.5} \\
\bottomrule
\end{tabular}

%% file: tables/table3_abl_train.tex
\begin{tabular}{@{}l cccc c cccc@{}}
\toprule
 & \multicolumn{4}{c}{Seen bodies} && \multicolumn{4}{c}{Unseen embodiments} \\
\cmidrule{2-5}\cmidrule{7-10}
 & SR$\uparrow$ & DE$\downarrow$ & CR$\uparrow$ & SPARC$\uparrow$ && SR$\uparrow$ & DE$\downarrow$ & CR$\uparrow$ & SPARC$\uparrow$ \\
\midrule
w/o adaptive gate & 90.0 & 0.50 & 73.0 & $-$2.02 && 87.5 & 0.64 & 61.8 & $-$2.05 \\
w/o style reweighting & 85.7 & 0.76 & 77.8 & $-$2.04 && 82.0 & 0.89 & 65.1 & $-$2.05 \\
\method & \textbf{93.0} & \textbf{0.38} & \textbf{79.6} & \textbf{$-$1.92} && \textbf{89.8} & \textbf{0.56} & \textbf{67.4} & \textbf{$-$1.97} \\
\bottomrule
\end{tabular}

%% file: tables/table4_abl_cond.tex
\begin{tabular}{@{}l cccc c cccc@{}}
\toprule
 & \multicolumn{4}{c}{Seen bodies} && \multicolumn{4}{c}{Unseen embodiments} \\
\cmidrule{2-5}\cmidrule{7-10}
 & SR$\uparrow$ & DE$\downarrow$ & CR$\uparrow$ & SPARC$\uparrow$ && SR$\uparrow$ & DE$\downarrow$ & CR$\uparrow$ & SPARC$\uparrow$ \\
\midrule
\multicolumn{10}{@{}l}{\textit{Body input}} \\
None & 80.5 & 1.02 & 66.0 & $-$2.12 && 76.5 & 1.18 & 55.5 & $-$2.19 \\
Body ID & 84.0 & 0.84 & 68.6 & $-$2.01 && -- & -- & -- & -- \\
\midrule
\multicolumn{10}{@{}l}{\textit{Where $z_e$ enters}} \\
w/o FiLM & 90.2 & 0.51 & 76.8 & $-$2.02 && 87.3 & 0.61 & 64.2 & $-$2.07 \\
w/o hyper-head & 87.5 & 0.62 & 74.5 & $-$2.05 && 84.0 & 0.75 & 62.0 & $-$2.10 \\
w/o teammate $z_e$ & 89.4 & 0.86 & 73.8 & $-$2.11 && 85.2 & 0.91 & 62.8 & $-$2.17 \\
\midrule
\method & \textbf{93.0} & \textbf{0.38} & \textbf{79.6} & \textbf{$-$1.92} && \textbf{89.8} & \textbf{0.56} & \textbf{67.4} & \textbf{$-$1.97} \\
\bottomrule
\end{tabular}

%% file: sections/5_conclusion.tex
\section{Conclusion}
\label{sec:conclusion}

We presented \method, a unified decentralised policy for physics-based cooperation among heterogeneous characters. Through a canonical body interface and cooperative control conditioned on each agent's and its teammates' capabilities, \method generalises across team compositions, unseen embodiments and objects, while acquiring new skills without compromising existing cooperation.

\noindent\textbf{Limitations.}
Although MorphHOI generalises across diverse embodiments and team compositions, coordination becomes more challenging as team size increases. Future work will explore more scalable cooperative control and richer interactions among heterogeneous characters.

%% file: sections/X_suppl.tex
\clearpage
\setcounter{page}{1}
\maketitlesupplementary
\appendix
\crefalias{section}{appendix}
\crefalias{subsection}{subappendix}
\crefname{appendix}{Appendix}{Appendices}
\Crefname{appendix}{Appendix}{Appendices}
\crefname{subappendix}{Appendix}{Appendices}
\Crefname{subappendix}{Appendix}{Appendices}

\noindent This supplementary material describes the embodiments and the evaluation protocol (\cref{sec:app_bodies}), the
implementation of \method, its rewards and its training (\cref{sec:app_impl,sec:app_rewards,sec:app_training}),
further analyses (\cref{sec:app_more}) of embodiment-conditioned reasoning (\cref{sec:app_ect}), the canonical action
representation (\cref{sec:app_iface}), the sensitivity to embodiment descriptions (\cref{sec:app_zswap}), scaling
embodiment diversity (\cref{sec:app_scale}), generalisation beyond training conditions (\cref{sec:app_sweep}) and the
baselines' failure modes (\cref{sec:app_compare}), and two qualitative
studies: teammates that join or leave a carry (\cref{sec:app_comego}) and the policy driving teams of robot models
(\cref{sec:app_robots}).

\section{Bodies and Evaluation Protocol}
\label{sec:app_bodies}

\subsection{Training Embodiments}
The policy trains on 320 simulated humanoids, mostly procedurally generated digital humans mixed with robot morphologies; the latter are simulated with their makers' kinematics, masses and joint limits and mapped to the canonical interface (\cref{sec:canonical}) like any other body. Each digital human is assembled from primitive shapes with sampled stature, body proportions, limb thickness, mass, and joint layout; between two canonical links, it may have zero to three hinges, in any order and with offsets or tilted axes. This procedure produces substantial variations in both body geometry and kinematic structure. Such diversity exposes the policy to different reaching capabilities, joint constraints, and physical properties during training. Consequently, agents must learn to coordinate across varying physical capabilities rather than relying on a fixed set of body configurations. Across all 320 bodies, statures range from 1.64 to 1.84\,m, masses from 46 to 80\,kg, and actuated DoFs from 18 to 28.

\subsection{Unseen Embodiments and Objects}
The unseen-embodiment setting evaluates generalisation to held-out embodiments with distinct body proportions, joint configurations, or PD gains outside the training range, none of which are encountered during training. These embodiments introduce variations in morphology and kinematic structure, testing whether the shared policy can coordinate with unfamiliar physical capabilities without further adaptation or task-specific retraining. The unseen-object setting evaluates generalisation beyond the training tables using novel object shapes: an ellipse, a hexagon, an octagon, a pentagon, a stadium, a rounded square, and a trapezoid. These shapes introduce different object geometries and grasping configurations, requiring agents to adapt their cooperative transport behaviours while maintaining coordinated lifting, carrying, and placement.

\subsection{Evaluation Settings}
\label{sec:app_eval}
Each evaluation episode runs the deterministic policy mean for 900 control steps (30\,s), without sampling from the action distribution. Agents are initialised at random positions up to 8\,m from the table, while the target is placed 3-10\,m away from it in a randomly sampled direction. These initial conditions create variations in approach distances, transport directions, and the coordination required throughout each episode. All methods are evaluated using the same simulator code and random seeds, ensuring identical initial spawns, target locations, and team compositions for fair comparisons. \Cref{tab:main} reports the mean performance over three evaluation seeds for each setting and team size. For the ablation studies, each metric is first averaged over three seeds and then aggregated across the two-, four-, and eight-agent settings using an unweighted arithmetic mean, so each team size contributes equally to the reported results.

\subsection{Metrics}
An episode is considered successful when the table reaches within 0.1\,m of the target, no agent falls, the table remains supported without dropping, and its final height is within 3\,cm of the resting height. These conditions jointly assess accurate transport, stable cooperation, and successful placement rather than merely reaching the target. The cooperation ratio (CR) measures the fraction of the carrying phase, from the initial lift until the table reaches the target, during which all agents maintain valid hand contacts within their embodiment-adaptive gates $\delta_e$ defined in \cref{eq:gate}. For box transport, contact is instead determined by whether both hands remain close to the box surface, reflecting the different grasping configuration. SPARC quantifies motion smoothness as the arc length of the normalised Fourier magnitude spectrum of the table's horizontal speed during transport, with a frequency cutoff of 10\,Hz. It is dimensionless and independent of transport distance and duration, allowing smoothness to be compared across different teams and trajectories. For each newly learned skill, SR$_\mathrm{H}$ is the harmonic mean of its success rate and the success rate of the same policy checkpoint on the original table task with the skill input disabled; CR$_\mathrm{H}$ is defined analogously. These harmonic-mean metrics evaluate whether new skills can be acquired while preserving the original cooperative behaviour, penalising substantial degradation in either task.

\section{Implementation Details}
\label{sec:app_impl}

\begin{table}[t]
  \centering
  \footnotesize
  \setlength{\tabcolsep}{3pt}
  \resizebox{\columnwidth}{!}{\input{tables/appendix_table1_hyper}}
  \caption{\textbf{Hyperparameters} and the symbols of the main paper: canonical interface
  (\cref{eq:ik,eq:reward,eq:gate}), network (\cref{eq:tokens,eq:blocks,eq:hyper}) and training recipe
  (\cref{eq:lora}); $\epsilon$ is the entropy weight of the contact assignment (\cref{eq:ot}) and $e$ the epoch.}
  \label{tab:hyper}
\end{table}

\begin{table}[t]
  \centering
  \footnotesize
  \setlength{\tabcolsep}{3pt}
  \resizebox{\columnwidth}{!}{\input{tables/appendix_table2_train}}
  \caption{\textbf{Optimisation, motion prior, simulation and evaluation.} $m$ and $h$ are a body's mass and pelvis
  height.}
  \label{tab:train}
\end{table}

\subsection{Observations}
Every component of an agent's observation is expressed in its own heading frame. The canonical state of
\cref{eq:canon_obs} has 226 dimensions: the root height and the position (15 links), 6-D rotation, linear and angular
velocity (15 links each), all in the body's own scale. The own token also
receives a short history of the agent's own motion and its row of the contact
assignment of \cref{eq:ot}. Each teammate contributes its angle around the table relative to the agent, its relative
planar position and 6-D relative rotation, a four-frame history (stride 2) of relative position, root height,
velocity, heading, hand heights, hand-rim distance and uprightness, and its row of the contact assignment.
The object is observed through its centre, 64 points on its rim, the rim points nearest to each hand and $M = 16$
contact points; the goal through the target position and a put-down flag. The body description
$z_e \in \mathbb{R}^{62}$ of \cref{eq:ze} holds six capability slots (log stature, arm and leg length ratios, mass
and joint strength relative to a human of that stature, one unused), the DoF mask $m_e \in \{0,1\}^{28}$
and the per-DoF log stiffness ratio.

\subsection{Network}
Tokens have $d=64$ channels, with two attention heads and feed-forward layers of width 512. The action head is an MLP with hidden sizes $[1024,512]$. Two embodiment-description encoders, shared across all bodies, provide complementary forms of conditioning. The first, $E$, is a two-layer MLP with 64 hidden and output dimensions, producing the description-dependent token offsets in \cref{eq:tokens}. The second, $\tilde E$, uses a 128-dimensional hidden layer and a 64-dimensional output to generate the conditioning features for FiLM. These features are projected through zero-initialised linear maps to obtain the scales and shifts of all six sublayers of each attention block and feed-forward sublayers, the embodiment-dependent action-head terms $A(\tilde z)$ and $c(\tilde z)$, and the modulation parameters for teammate tokens. The low-rank action-head projection uses $B \in \mathbb{R}^{8 \times 512}$, allowing body-specific adjustments while retaining a shared action mapping. Contact assignment in \cref{eq:ot} is approximated using ten Sinkhorn iterations with entropy regularisation $\epsilon=0.3$, and the task-attention bias coefficient $\beta_h$ is initialised at 1. The team-attention bias incorporates pairwise distance, closing speed, and the overlap between agents' contact-assignment rows. These relational features allow attention to account for both the agents' relative motion and whether they are expected to interact with similar regions of the object.

\subsection{Contact Assignment}
\label{sec:app_ot}
The assignment $\hat P \in \mathbb{R}^{N \times M}$ of \cref{sec:ect} is computed once per control step from the planar positions $u_j$ of the team's $N$ agents $j$ (including $i$) and $p_m$ of the $M$ contacts. It is the solution of entropic optimal transport~\cite{cuturi2013sinkhorn} between agents and contacts,
\begin{equation}
\begin{aligned}
  &P = \textstyle\arg\max_{P \in U(\kappa, \nu)} \sum_{j,m} P_{j,m} S_{j,m} + \epsilon H(P), \\
  &S_{j,m} = -\alpha \lVert p_m - u_j \rVert + h_\psi(p_m - u_j,\, \dot u_j,\, z^j),
\end{aligned}
  \label{eq:ot}
\end{equation}
where $H$ is the entropy, $\alpha > 0$ a learned scale, $\dot u_j$ the agent's planar velocity, and $U(\kappa,\nu)$ the couplings whose rows sum to the agent capacities $\kappa_j$ and whose columns sum to the contact importances $\nu_m$. The learned score $h_\psi$ adds two terms to the distance: an MLP of the approach geometry, which holds the offset $p_m - u_j$ along the contact's inward normal and tangent, its length and the agent's closing speed towards the contact, and a bilinear affinity between the description $z^j$ and the position of the contact on the object. An MLP predicts the capacity $\kappa_j \in (0, 2)$ from $z^j$, and the importances $\nu_m$ are a softmax over the contacts of an MLP of each contact's position on the object, uniform at initialisation and scaled to the total capacity $\sum_j \kappa_j$. We run 10 Sinkhorn iterations with $\epsilon = 0.3$, ending on the row update, and normalise each row, $\hat P_{j,m} = P_{j,m} / \kappa_j$, so that $\hat P_j$ is a distribution over the contacts. All terms of $S$ are invariant to the observer's frame. The assignment enters the policy three times: each agent's row is embedded into its token (\cref{eq:tokens}) as its expected contact, the $\hat P_j$-weighted mean of the contacts' positions, normals and tokens; the overlap $\sum_m (\hat P_{i,m} \hat P_{j,m})^{1/2}$ of two rows is a feature of $B^\mathrm{rel}$; and the row of agent $i$ gives the bias $B^\mathrm{ot}$ of task attention (\cref{eq:bias}).

\subsection{Canonical Interface and Motion Prior}
\Cref{eq:ik} runs six damped Gauss-Newton iterations per control step with $\lambda = 0.02$, steps clipped to
0.5\,rad and targets rate-limited to 0.5\,rad per step. One
discriminator serves all bodies. It reads five consecutive frames of 125 canonical features, scaled by the ratio of the
reference pelvis height to the body's, and velocities by the square root of that ratio, so that a
smaller body walking at the same relative pace produces the same features as the reference human. Two copies are
trained, a full one on twelve walking and turning clips and an arm-masked one, which ignores the arm features, on 24
clips that add three box pick-ups and the walking clips played in reverse; the style reward blends them as described in
\cref{sec:app_rewards}. The critic is a six-layer transformer that alternates self- and cross-attention. PD gains come from each agent model's joint groups.

\input{sections/X_rewards}

\section{Further Analyses}
\label{sec:app_more}

\begin{table}[t]
  \centering
  \footnotesize
  \setlength{\tabcolsep}{1.6pt}
  \resizebox{\columnwidth}{!}{\input{tables/appendix_table5_abl_ect}}
  \caption{\textbf{Inside ECT}, metrics and team sets as in \cref{tab:abl_cond}. Every variant is retrained with the
  full recipe.}
  \label{tab:abl_ect}
\end{table}

\begin{table}[t]
  \centering
  \footnotesize
  \setlength{\tabcolsep}{1.6pt}
  \resizebox{\columnwidth}{!}{\input{tables/appendix_table6_abl_iface}}
  \caption{\textbf{Action space}, metrics and team sets as in \cref{tab:abl_cond}. Every variant is retrained with
  the full recipe.}
  \label{tab:abl_iface}
\end{table}

\begin{figure}[t]
  \centering
  \includegraphics[width=\columnwidth]{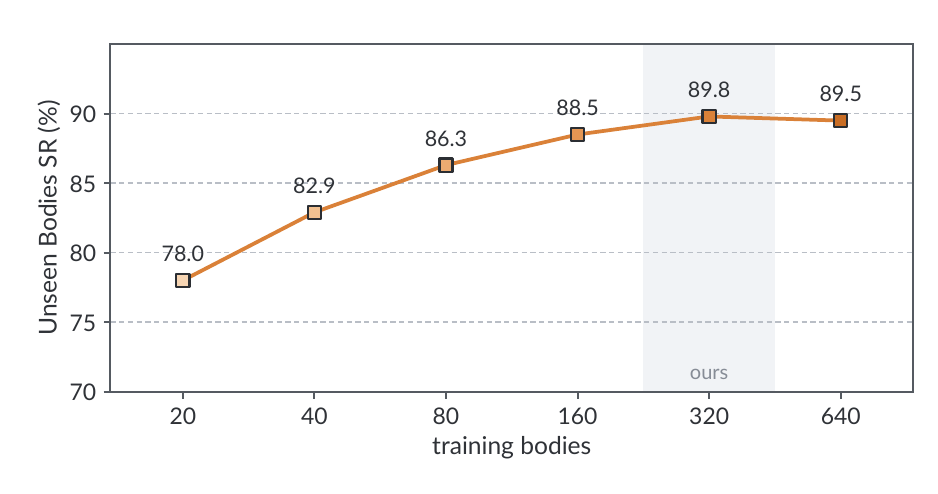}
  \vspace{-2.4em}
  \caption{\textbf{More training bodies}, unseen embodiments, two to eight agents: SR against the number
  of training bodies $K$. Shaded:
  the set used in the main paper.}
  \label{fig:trainbodies}
  \vspace{-0.8em}
\end{figure}

\begin{figure*}[t]
  \centering
  \includegraphics[width=\textwidth]{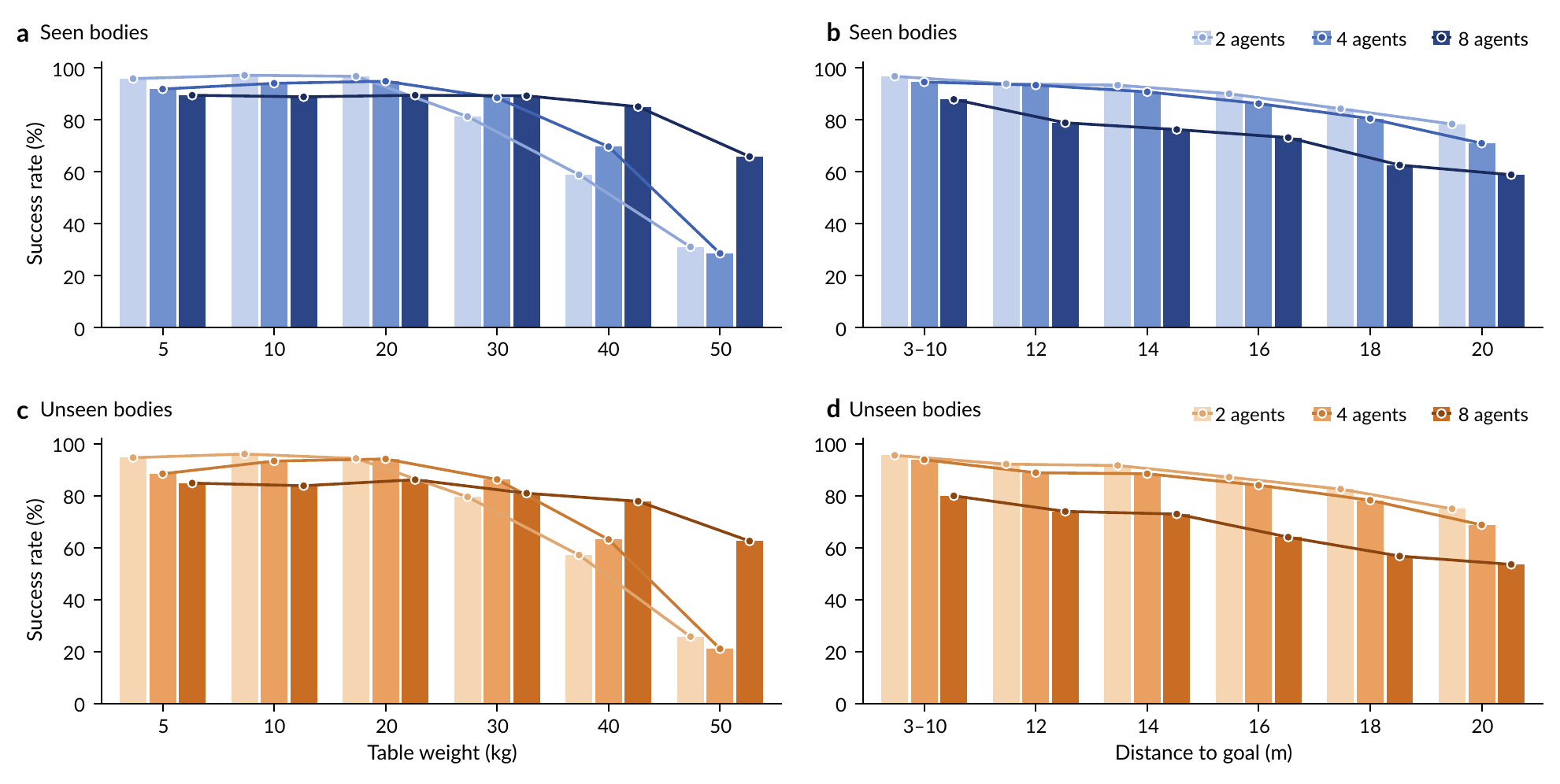}
  \vspace{-2.0em}
  \caption{\textbf{Generalisation to heavier tables and farther goals.} Success rates of the Stage-II policy on seen and unseen embodiments with teams of two, four, and eight agents, evaluated under increasing table masses (a, c) and goal distances (b, d).}
  \label{fig:sweep}
  \vspace{-0.8em}
\end{figure*}

\subsection{Embodiment-Conditioned Reasoning}
\label{sec:app_ect}

\Cref{tab:abl_ect} examines the role of structured attention in ECT. \underline{Flat attention} replaces the separate self-, team-, and task-attention operations with a TeamHOI-style stack while retaining both biases, FiLM, and the generated head. \underline{W/o OT task bias} removes the assignment bias from task attention, while \underline{w/o relational bias} removes the pairwise bias from team attention. Removing either bias reduces SR from 93.0\% to 84.2\% or 87.4\% and increases DE from 0.38 to 0.71 or 0.57, despite relatively small changes in CR; the gap widens on unseen embodiments (89.8\% versus 80.0\% or 82.1\%). This reveals that maintaining contact is not sufficient for effective cooperation: agents must also agree on where to grasp and how to coordinate their movements towards the target. Separating the three attention operations further improves SR over flat attention (93.0\% versus 90.4\%) and produces smoother transport, suggesting that explicitly modelling team relationships and task constraints benefits coordination beyond the structural biases alone.

\subsection{Canonical Action Representation}
\label{sec:app_iface}

\Cref{tab:abl_iface} compares two alternatives to the canonical joint actions of \cref{eq:ik}, keeping the network, PD scaling, body descriptions and DoF masks unchanged. \underline{Hinges by name} directly maps canonical targets to similarly named hinges without IK, while \underline{Links + IK} commands link orientations in the pelvis frame and solves them onto the body's hinges, aligning the action representation with the link-level observations. Without IK, the same action produces different rotations across embodiments, reducing SR from 93.0\% to 35.1\%. Although link-level actions align with observations, they couple each command to multiple upstream joints, reaching only 82.0\% SR. Canonical joint targets instead provide local, embodiment-consistent control aligned with the motion prior, achieving 93.0\% SR. This suggests that observation and action representations need not be symmetric: link-level states support shared perception, while joint-level targets enable more precise control.

\begin{table}[t]
  \centering
  \footnotesize
  \setlength{\tabcolsep}{1.6pt}
  \resizebox{\columnwidth}{!}{\input{tables/appendix_table7_abl_zswap}}
  \caption{\textbf{Wrong descriptions at test time.} The stage-II policy given each description, metrics and team
  sets as in \cref{tab:abl_cond}.}
  \label{tab:abl_zswap}
  \vspace{-1.6em}
\end{table}

\IfFileExists{figures/appendix_fig4_compare_analysis.pdf}{%
\begin{figure*}[t]
  \centering
  \includegraphics[width=\textwidth]{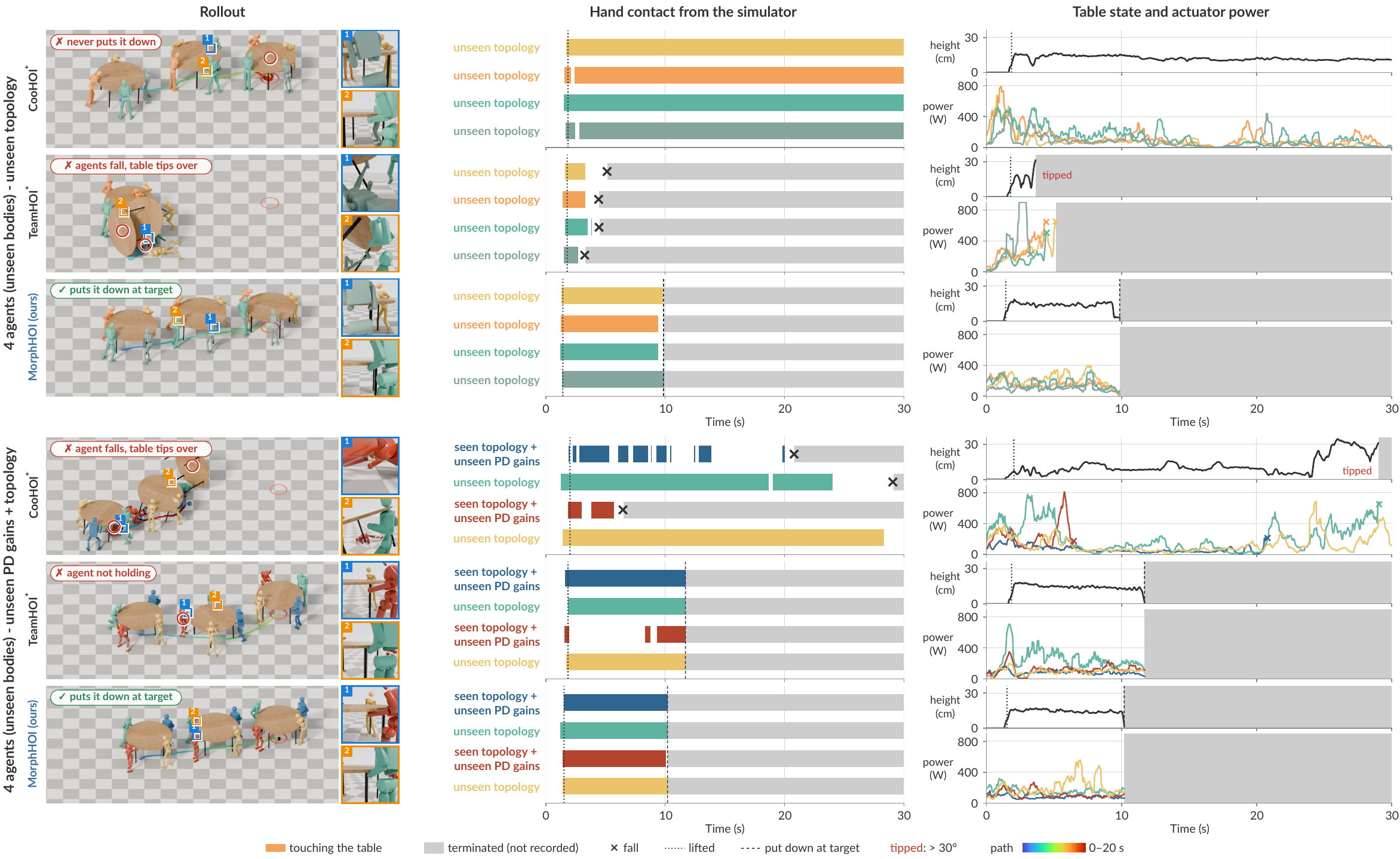}
  \vspace{-1.6em}
  \caption{\textbf{Why the baselines fail on unseen embodiments}, under identical teams, spawns and targets. Top: four
  unseen bodies, all with unseen topologies. Bottom: four unseen bodies, two with unseen PD gains (on seen topologies)
  and two with unseen topologies. For each method: the rollout with close-ups (1, 2) of hand-object contact and the
  object path coloured by time; the hand contact of each agent as reported by the simulator; and the table height and
  the actuator power of each agent. Dotted line: lift; dashed line: put-down at the target; cross: an agent falls; grey:
  the episode has ended; tipped: table tilt above $30^\circ$.}
  \label{fig:compare_analysis}
  \vspace{-0.8em}
\end{figure*}}{}

\subsection{Sensitivity to Embodiment Descriptions}
\label{sec:app_zswap}

\Cref{tab:abl_zswap} tests the sensitivity of ECT to embodiment descriptions while retaining each body's correct DoF mask. \underline{Teammate's} assigns each agent the next teammate's description, preserving the team's overall capabilities but mismatching their identities; \underline{farthest body's} substitutes the most dissimilar training body. We also perturb capability features and stiffness ratios with Gaussian noise scaled by their variation across training bodies. Swapping descriptions reduces SR from 93.0\% to 71.2\%, and further to 66.2\% for the farthest body, showing that cooperation depends on correctly associating capabilities with each participant rather than merely knowing the team's overall capabilities. In contrast, Gaussian noise causes limited degradation even at $\sigma=1$. Interestingly, moderate noise ($\sigma=0.5$) slightly improves SR on unseen embodiments (90.3\% versus 89.8\%), suggesting tolerance to parameter uncertainty and a possible smoothing effect reminiscent of domain randomisation. Together, these results indicate that ECT is sensitive to incorrect capability attribution but robust to moderate inaccuracies in physical descriptions.

\subsection{Scaling Embodiment Diversity}
\label{sec:app_scale}

\Cref{fig:trainbodies} examines how training embodiment diversity affects generalisation. We train on nested sets of $K \in \{20, 40, 80, 160, 320, 640\}$ bodies with the same training schedule, and evaluate on unseen embodiments across teams of two, four and eight. SR increases from 78.0\% with 20 bodies to 86.3\% with 80 and 89.8\% with 320. This suggests that greater embodiment diversity helps the policy learn transferable cooperative behaviours rather than overfitting to particular body combinations. However, performance saturates beyond 320 bodies, reaching 89.5\% with 640. We hypothesise that this saturation partly reflects the fixed training budget, which limits exposure to individual embodiments and team compositions as diversity increases. We therefore use 320 bodies in the main experiments.

\subsection{Generalisation Beyond Training Conditions}
\label{sec:app_sweep}

\Cref{fig:sweep} evaluates the stage-II policy beyond its training range in table mass and target distance. In panel (a), all teams carry tables of equal mass, revealing how team size affects load tolerance. As the load increases, success rates generally decline, but eight-agent teams retain substantially higher robustness under heavy loads, benefiting from distributed load sharing. Although four-agent teams occasionally underperform pairs at extreme loads, the overall results suggest that larger teams can better tolerate heavier objects. Panel (b) reveals the opposite trend for longer carries: smaller teams maintain higher success rates, while larger teams become increasingly vulnerable to accumulated coordination failures. Notably, seen and unseen embodiments exhibit similar trends across both sweeps, suggesting consistent generalisation across task demands and previously unseen bodies. Together, these results highlight a trade-off: larger teams generally offer greater load tolerance, but sustaining coordination becomes more challenging as transport distances increase.

\IfFileExists{figures/appendix_fig5_comego.pdf}{%
\begin{figure*}[t]
  \centering
  \includegraphics[width=\textwidth]{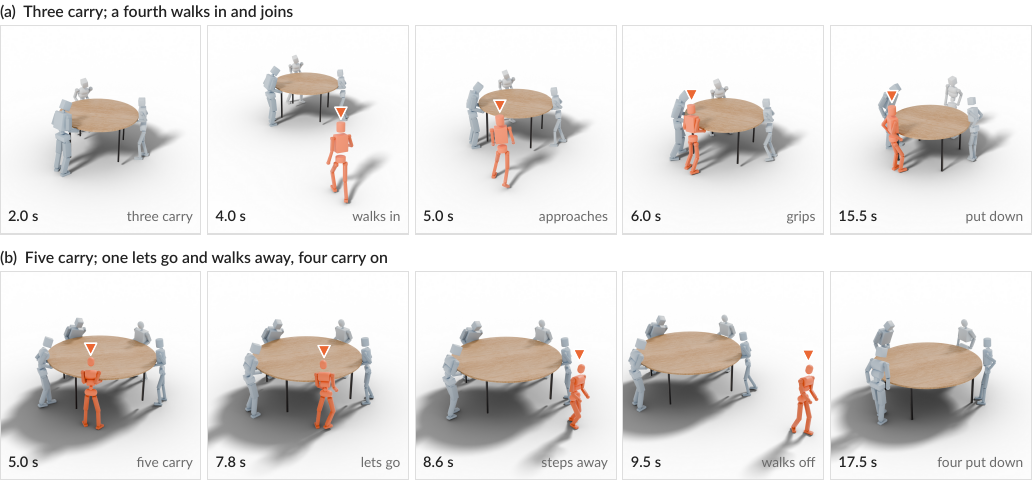}
  \caption{\textbf{Teammates come and go}, the stage-II policy unchanged. (a) Three synthetic bodies carry a 1.0\,m
  round table; a fourth (orange) enters at 3.0\,s, 5\,m from the table, walks in, grips the rim at about 6\,s, and the
  four put the table down at about 15\,s. (b) Five bodies carry a 1.4\,m round table; at 6.7\,s one of them (orange)
  leaves the team, lets go, steps back and walks off, and the remaining four carry the table on and put it down at
  about 17\,s.}
  \label{fig:comego}
  \vspace{-0.8em}
\end{figure*}}{}

\subsection{Failure Modes of the Baselines}
\label{sec:app_compare}

\Cref{fig:compare_analysis} supplements the unseen-embodiment settings of \cref{fig:compare} through the simulator's
contact and actuator signals. With four unseen topologies (top), CooHOI$^*$ lifts the table and all agents keep their
grip, yet the team never puts it down: the table stays in the air for the full 30\,s. TeamHOI$^*$ loses contact within
a few seconds of the lift, all four agents fall, and the table tips over. \method keeps every agent in contact from the
lift until the put-down at the target after about 10\,s, with lower actuator power than CooHOI$^*$ throughout. When two
teammates also have unseen PD gains (bottom), the baselines fail first on exactly these agents. Under CooHOI$^*$, both
lose and regain contact repeatedly and eventually fall, after which the table tips over. Under TeamHOI$^*$, the team
reaches the target, but one of them holds the rim only briefly, leaving the load to the other three. With \method, all
four agents, including those with unseen gains, hold the table continuously and put it down at the target. Since the
baselines share our canonical observations and actions, these failures indicate that a shared interface alone does not
convey how differently each body responds to the same command; conditioning on the body descriptions, including the
per-DoF stiffness ratios, lets \method adapt its control and its cooperation to each teammate.

\section{Teammates Come and Go}
\label{sec:app_comego}

In training, a team is drawn at every reset and keeps its members for the whole episode. At run time, however, every
agent reads its teammates' tokens and descriptions anew at every control step, so the team the policy cooperates with
is whatever team is present at that moment. \Cref{fig:comego} changes the team during a carry without changing the
policy. In (a) a fourth agent is activated beside a team of three already carrying the table: it walks in, finds a free
place on the rim through the contact assignment of \cref{eq:ot}, which now includes it, and grips, while the three
carriers re-space around the table to make room, and the four put the table down together. In (b) one of five carriers
is removed from the team mid-carry: its token disappears from the others' team attention and from their grip gate, the
assignment redistributes its contacts among the remaining four, and they re-space and carry the table on to the target,
while the departing agent, driven by the same policy towards a goal away from the team, lets go and walks off. No agent
falls in either episode. The behaviour follows from a policy whose team is an input rather than a fixed part of the
network: the same mechanism that lets one policy serve teams of any composition lets it absorb a change of
composition while the object is in the air.

\section{Potential Contributions to Robotics}
\label{sec:app_robots}

\method offers three potential benefits for robotics. First, it controls heterogeneous robot teams in simulation with a single cooperative policy, without embodiment- or team-specific controllers, suggesting a foundation for cross-platform cooperation (\cref{sec:app_found}). Second, the policy could serve as a transferable teacher, enabling robot-specific controllers to acquire cooperative behaviours through initialisation or distillation rather than training from scratch (\cref{sec:app_teacher}). Third, \method generates physically simulated multi-robot interaction data, including coordinated motions, contact forces, and object dynamics. These data could support imitation learning, motion priors, and studies of heterogeneous load sharing (\cref{sec:app_data}).

\subsection{A Cooperative Foundation Policy}
\label{sec:app_found}

Humanoid robots provide a challenging test of unified embodiment control, as their joint structures, mass distributions, and actuator properties differ substantially from those of the reference human. As shown in \cref{fig:robots}, the same stage-II policy controls teams of two, four, and eight robots carrying tables. The learned skills also apply to robot-only teams: robot pairs carry a long box or descend more steps than encountered during training, while four robots reorient a table to navigate a narrow passage using an additional pose-conditioned skill. These results demonstrate that the same canonical interface enables a single policy to control both digital humans and robot models without robot-specific controllers. Rather than requiring a separate controller for each robot type or team composition, \method conditions cooperation on the physical capabilities of all participants. This suggests its potential as a reusable foundation for heterogeneous multi-robot interaction, supporting diverse embodiments through their descriptions and new tasks through lightweight skill adaptation.

\subsection{A Teacher for Robot-Specific Controllers}
\label{sec:app_teacher}

Although a shared policy enables cooperation across different embodiments, practical deployment may benefit from smaller, platform-specific controllers tailored to hardware constraints. \method could serve as a pretrained teacher for such controllers, transferring learned cooperative behaviours through policy distillation or providing an initialisation for further adaptation~\cite{rusu2015policy,schmitt2018kickstarting,ross2011reduction}. Its canonical actions can be mapped to each robot's native joints through \cref{eq:ik}, providing embodiment-compatible control targets. Rather than learning locomotion, object manipulation, and multi-agent coordination independently, a specialised controller could build upon the cooperative behaviours already acquired by the shared policy. This offers a potential route towards efficient robot-specific control while retaining the benefits of heterogeneous cooperative learning.

\subsection{Physically Simulated Cooperative HOI Data}
\label{sec:app_data}

Cooperative interaction data involving multiple robots remain scarce, while human motion captures require retargeting and do not directly provide the contact forces or object dynamics underlying the interactions~\cite{araujo2026retargeting}. \method offers an alternative by generating physically simulated cooperative motion directly on robot models. We record teams of two, three, four, and eight robots, together with box-carrying, stair-descent, and pose-conditioned transport skills. The framework can generate approximately 120 minutes of team motion (8.0 robot-hours) in five minutes on a laptop GPU (RTX 3080 Ti), recording link poses, joint states and torques, contact forces, and object trajectories at 30\,Hz. As illustrated in \cref{fig:robodata}, these recordings capture not only coordinated movements but also how robots establish contact, distribute loads, and expend energy during transport. Within the simulator's dynamics and actuator models, such data provide physically grounded interaction trajectories beyond conventional kinematic motion captures. They could support imitation learning, motion-prior training, and specialised multi-robot control, while enabling systematic studies of heterogeneous load sharing and contact coordination.

\subsection{Future Real-World Deployment}
An important next step is to deploy \method on physical humanoid robots to evaluate how the learned cooperative behaviours transfer to real-world interactions. Such deployment will introduce additional challenges, including actuator limitations, imperfect state estimation, communication delays, and discrepancies between simulated and physical contact dynamics. Addressing these challenges will be essential for translating embodiment-aware cooperation into practical robotic systems. The present work establishes and systematically evaluates a unified policy for heterogeneous cooperative interaction, providing a foundation for investigating how shared cooperative behaviours can be preserved under real-world physical and sensing constraints. Future hardware deployment will therefore serve not only to validate the framework but also to reveal the adaptations needed for robust cooperation among physical robots.

\IfFileExists{figures/appendix_fig6_robots.pdf}{%
\begin{figure*}[p]
  \centering
  \includegraphics[width=\textwidth]{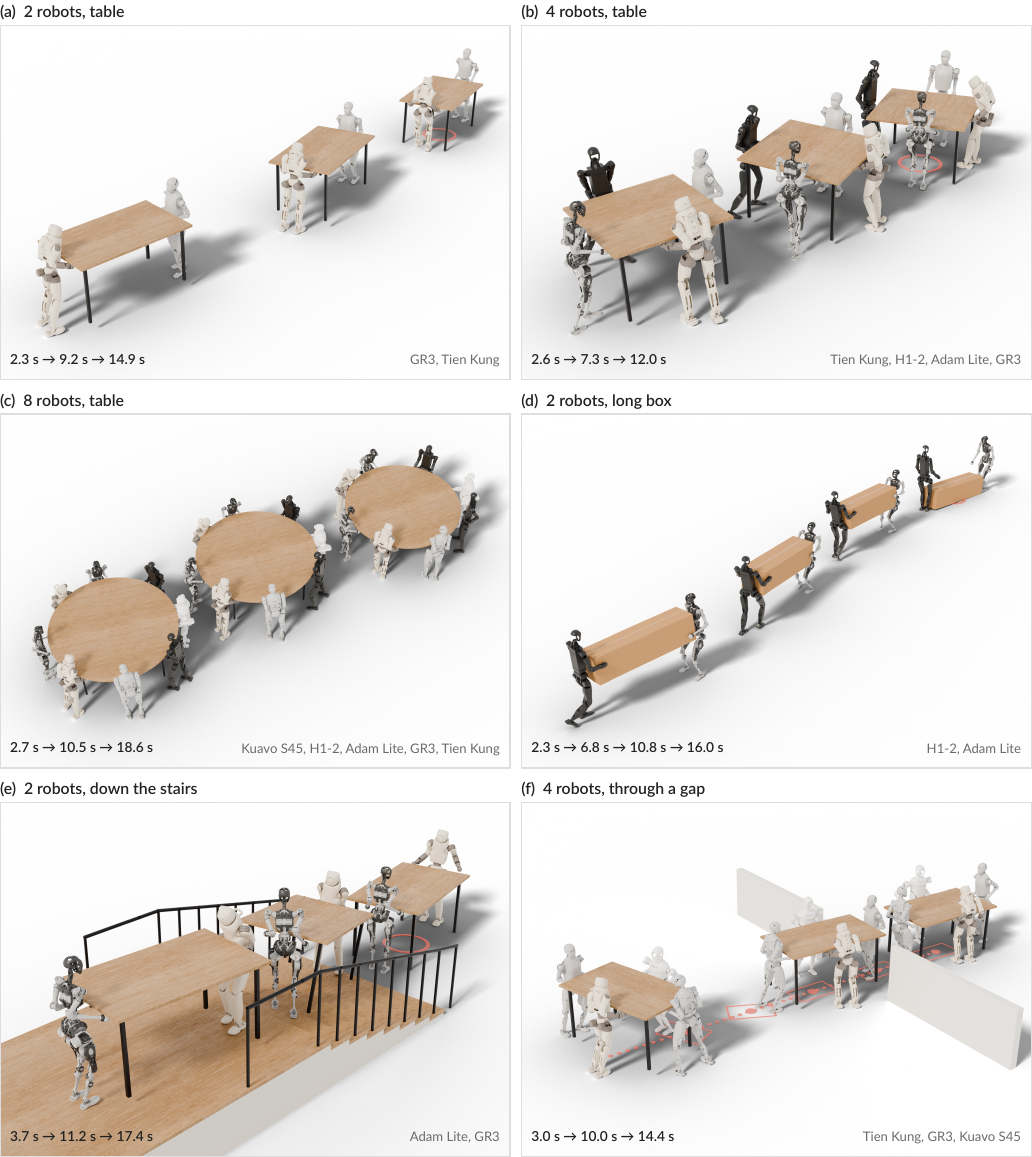}
  \caption{\textbf{Teams of robot models}, rendered with the makers' meshes and colours; each tile shows the team at
  the listed times. (a-c) The stage-II policy without any skill: two robots with a $2.0 \times 1.2$\,m table, four
  robots, one per side, with a $1.6 \times 1.6$\,m table, and eight robots with a round table of 2.8\,m diameter.
  (d) Two robots with a $2.0 \times 0.5 \times 0.5$\,m box, with the box skill of \cref{sec:exp_what}. (e) Two robots
  carrying a table down eight steps of 8\,cm with the down-stairs skill of \cref{sec:exp_what}, unchanged. (f) Four
  robots turning a table from crosswise to lengthwise through a 2.2\,m gap, with the pose skill, which adds a target heading
  to every waypoint of a path; red: the commanded path and table poses.}
  \label{fig:robots}
\end{figure*}}{}

\IfFileExists{figures/appendix_fig7_robodata.pdf}{%
\begin{figure*}[p]
  \centering
  \includegraphics[width=\textwidth]{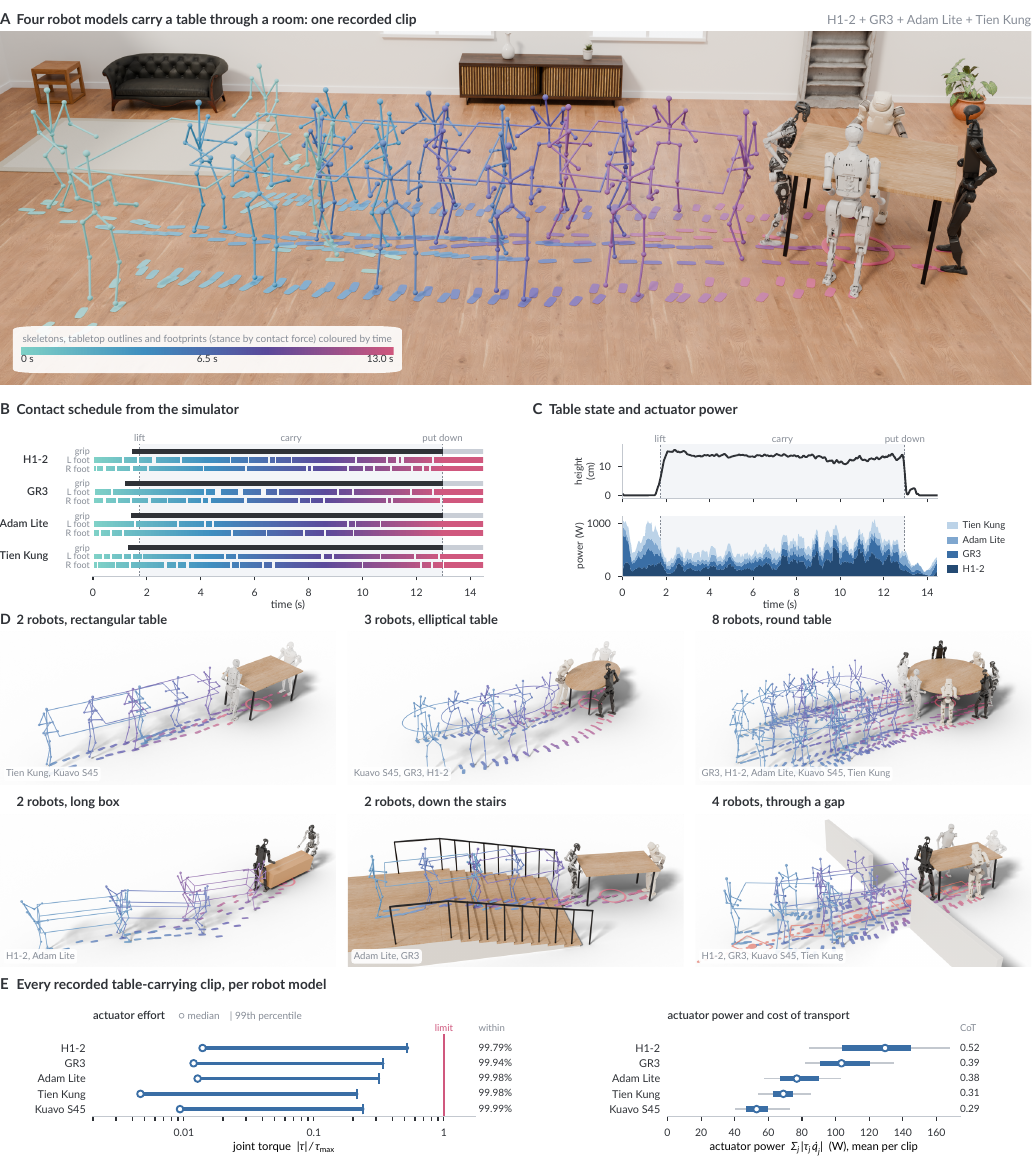}
  \caption{\textbf{Physically simulated motion data of robot teams} from \method: the stage-II policy unchanged
  (A-C, D top) and with the skills of \cref{fig:robots} (D bottom). (A) One clip:
  joint skeletons and tabletop outlines at earlier times and footprints where the simulator reports a loaded foot,
  coloured by time; the robots at the put-down. (B) Contact schedule from the simulator's contact forces (grip: an
  arm link on the table, brief touches removed; stance: a foot carrying at least 30\% of its robot's weight).
  (C) Table height and stacked actuator power $\sum_j |\tau_j \dot q_j|$ (0.3\,s average). (D) Further clips.
  (E) Over all successful table clips: torque relative to the rated effort and the share of samples within it; mean
  actuator power per clip (box: quartiles, line: 5th-95th percentile) and median cost of transport
  $\int \sum_j |\tau_j \dot q_j|\,dt \,/\, (m g d)$, $d$ the distance walked. Every pose is a recorded frame.}
  \label{fig:robodata}
\end{figure*}}{}

%% file: tables/appendix_table1_hyper.tex
\begin{tabular}[t]{@{}l l l@{}}
\toprule
Symbol & Meaning & Value \\
\midrule
\multicolumn{3}{@{}l}{\textit{Canonical interface}} \\
$\lambda$ & IK damping, iterations & 0.02, 6 \\
-- & IK step clip, target rate limit & 0.5\,rad, 0.5\,rad/step \\
$w_G$, $w_S$ & task / style weight & 0.5, 0.5 \\
$\bar w$ & style weight of non-reference bodies & 0.3 \\
$\delta_0$, $\Delta h_\mathrm{max}$ & contact gate & 0.04\,m, 0.06\,m \\
$k_z$ & put-down sharpness & 10\,m$^{-1}$ \\
\midrule
\multicolumn{3}{@{}l}{\textit{Network}} \\
$d$, heads & token width, attention heads & 64, 2 \\
$L$ & blocks of self / team / task attention & 2 \\
-- & feed-forward width & 512 \\
-- & tokenizer / action head MLP & $[256, 128]$ / $[1024, 512]$ \\
$M$ & contact points, rim points & 16, 64 \\
$\epsilon$ & assignment entropy, iterations & 0.3, 10 \\
$r$ & rank of the head term & 8 \\
\midrule
\multicolumn{3}{@{}l}{\textit{Training recipe}} \\
-- & stage I / stage II epochs & 9250 / 2600 \\
$\rho_\mathrm{mix}$ & share of mixed teams in stage II & 0.7 \\
$r_s$ & LoRA rank, body / head & 8 / 16 (box 32 / 32) \\
-- & skill epochs: path / pose / box & 2600 / 3000 / 10000 \\
-- & skill epochs: stairs down / up & 7000 / 7000 \\
\bottomrule
\end{tabular}

%% file: tables/appendix_table2_train.tex
\begin{tabular}[t]{@{}l l@{}}
\toprule
Parameter & Value \\
\midrule
\multicolumn{2}{@{}l}{\textit{PPO}} \\
optimiser, learning rate & Adam, $2 \times 10^{-5}$ \\
discount $\gamma$, GAE $\lambda$ & 0.99, 0.95 \\
clip ratio & 0.2 \\
horizon, mini-epochs & 32 steps, 6 \\
minibatch per GPU & 4096 agent samples \\
value / bound / entropy coefficient & 5 / 10 / 0 \\
action std & 0.055 \\
\midrule
\multicolumn{2}{@{}l}{\textit{Discriminator}} \\
input & 5 frames $\times$ 125 features \\
MLP & $[1024, 512]$ \\
gradient penalty / logit reg. / weight decay & 5 / 0.01 / $10^{-4}$ \\
loss coefficient & 5 \\
demo / agent minibatch & 512 / 1024 \\
replay buffer, keep probability & $2 \times 10^5$, 0.01 \\
motion clips: full / arm-masked & 12 / 24 \\
\midrule
\multicolumn{2}{@{}l}{\textit{Simulation}} \\
simulator & Isaac Gym~\cite{makoviychuk2021isaac} \\
simulation / control rate & 60\,Hz (2 substeps) / 30\,Hz \\
GPUs $\times$ environments $\times$ slots & 4 $\times$ 256 $\times$ 4 \\
PD gains, synthetic bodies & $k_p \propto m h$, $k_d \propto m h^{1.5}$, $k_d \ge 0.1 k_p$ \\
\midrule
\multicolumn{2}{@{}l}{\textit{Evaluation}} \\
episode & 30\,s (900 steps) \\
success: distance / height above rest & 0.1\,m / 3\,cm \\
lifted (start of the carry) & 7\,cm \\
spawn / target distance & $\le 8$\,m / 3--10\,m \\
\bottomrule
\end{tabular}

%% file: sections/X_rewards.tex
\section{Rewards}
\label{sec:app_rewards}

\begin{table*}[t]
  \centering
  \footnotesize
  \setlength{\tabcolsep}{9pt}
  \input{tables/appendix_table3_rewards}
  \caption{\textbf{Table-task reward.} Each agent receives rewards per
  control step (30\,Hz); the columns give the weights in the two reward phases of stage I, the second beginning once the reward of the first
  levels off (epoch 6400 for \method), and in stage II; grey:
  inactive. $d_\mathrm{nn}$: distance of the root to the nearest rim
  point, $v_n$: root speed towards it along the rim normal $n$; walk\_pos and walk\_vel are 1 within 0.2\,m of the
  standing distance of 0.3\,m, and walk\_pos also while $G = 1$. $f$: facing of the agent; $u_c$, $u_g$: directions to
  the table centre and from the table to the target; $\sigma_d = \sigma(10(d_\mathrm{nn} - 1))$.
  $r_\mathrm{ang} = \exp(-3\, \overline{(\Delta\theta - 2\pi/k)^2})$ scores the angular spacing of the $k$ agents around
  the table and $r_\mathrm{cov}$ the share of the table's extent covered by their rim points. $d_h$: distance of hand
  $h$ to the rim; $\phi_h$: placement factor (hand above the rim, hands apart and level); $\mathcal{C}$: hands within the
  contact gate $\delta_e$ of \cref{eq:gate}; $a_e = \max(0.10, \delta_e + 0.02)$; $\rho_h$: rise of the rim at hand
  $h$ towards the carry height $z^*$. $G$: 1 while every agent has both hands in $\mathcal{C}$. $x_o$, $x_g$: planar
  positions of the table and the target; $\dot d$: rate at which their distance shrinks; $f_\mathrm{far}$: facing of
  the agent farthest from the target; align is 1 within 0.5\,m. $P$: put-down mode, latched once
  $\lVert x_g - x_o \rVert < 0.03$\,m; under $P$, walk\_face, formation, hand, contact, lift, align, progress and
  transport are held at 1, and in the last phase transport is replaced by stay. The hand factor of putdown is 1 once
  both hands are 7\,cm off the rim. $z_o$, $z_o^\mathrm{rest}$: table height and its rest height; $D$: output of the
  discriminators. Heights $z^* = 0.94\,s$ and $z_\mathrm{put} = 0.65\,s$ scale with the table height through $s$.
  Lengths in m, speeds in m/s.}
  \label{tab:rewards}
\end{table*}

\begin{table}[t]
  \centering
  \footnotesize
  \setlength{\tabcolsep}{3pt}
  \resizebox{\columnwidth}{!}{\input{tables/appendix_table4_skill_rewards}}
  \caption{\textbf{Rewards of the new tasks.} Path, pose and stairs keep the terms of \cref{tab:rewards} (last column)
  and change those listed, pose also taking the changes of path; the box reward replaces it. Path: transport, align and progress aim at the current waypoint
  $x_j$; $e_\mathrm{seg}$: distance of the table to the segment from the previous waypoint to $x_j$. Pose: $e_\psi$,
  wrapped error of the table's heading to the target heading at $x_j$; $k$: team size. Stairs: heights of hands and rim
  points are taken above the ground below them. Box: $p_g$, the agent's assigned grasp point at the centre of an end of the box,
  0.25\,m in from that end face, at height $z_g$; $p_s$, the standing point 0.2\,m beyond that face; $u_p$: direction to
  $p_g$; $\bar p_\mathrm{hand}$: mean of the hands; $\mathbf{x}$: 3-D positions; $a_o$, $a_g$: axes of the box and the
  target pose. $L$: 1 while the box centre is at least 0.6\,m high (carry\_far, carry\_near: or within 0.5\,m of the
  target, where carry\_far is 1); $N$: 1 within 0.5\,m of the target; walk\_pos is 1 within 0.2\,m of $p_s$.}
  \label{tab:skill_rewards}
\end{table}

\subsection{Structure}
\Cref{tab:rewards} lists every term of the table task. The split between task and style rewards and the approach,
formation, hand, transport, alignment and put-down terms follow TeamHOI~\cite{lionar2026teamhoi}, including the gating
of the facing, hand and lift terms by the formation scores $r_\mathrm{ang}$ and $r_\mathrm{cov}$. The style reward blends the full discriminator with the arm-masked one through the weight
$\sigma(15\,(1.0\,\mathrm{m} - d_\mathrm{hand\text{-}rim}))$, so that the arms are scored by the motion prior while
walking and are left to the task reward once the hands approach the rim. The changes concern what heterogeneous bodies
and long carries require: a contact gate per body, a lift term paid by the rise of the table at twice the weight, a
signed progress term, and two terms that bring the table to rest at the target. All heights, including the upper bound
of the table that ends an episode, scale with the table height through $s$.

\subsection{Approach, Contact and Lift}
Before the team holds the table, the walking terms draw every agent to a standing distance of 0.3\,m from its nearest
rim point at a walking speed of 1.5-2.5\,m/s, facing the rim when close and the table centre when far, and the
formation term spreads the team over the rim. The contact term is where bodies differ. A body whose wrists hang
$\Delta h_e = \max(0, h_\mathrm{table} - h_{\mathrm{wrist},e})$ below the rim in its standing pose has the contact gate
$\delta_e = 0.04 + \min(0.06, \Delta h_e)$\,m: a hand counts as in contact within $\delta_e$ of the rim, and the contact
reward grows over the radius $a_e$, at least 0.10\,m. A short body is thus paid for the grip it can achieve. The lift term pays each hand in contact by how far it has raised the rim from its rest height
towards the carry height, $\rho_h = \mathrm{clip}((z_h - z_\mathrm{rest})/(z^* - z_\mathrm{rest}), 0, 1)$, and
penalises lifting above $z^*$. Unlike TeamHOI's term, which pays the distance to the carry height, it rewards every
centimetre of rise, which matters for bodies that cannot reach the carry height in one motion.

\subsection{Carry and Arrival}
The grip gate $G$ makes the transport, alignment and progress terms pay only while every member of the team holds the
table. The signed progress term pays the speed of the table towards the target, normalised by 0.6\,m/s; it is one-sided
at weight 0.5 for the first 6400 epochs, while the team learns to carry, and signed at weight 1 afterwards, so that
moving the table away from the target is penalised. Once the table comes within 0.03 m of the target, the put-down mode $P$ latches for the rest of the episode: the put-down flag is activated, and the facing, formation, hand, grip and carry terms are held at 1, and the put-down term pays hands near the put-down height or released from the rim
together with a standing root. From epoch 6401, the stay term keeps the table at the target without the grip gate, and
the lowering term pays the table's return to its rest height, with episodes extended from 600 to 750 steps to leave
time for it. The lowering term is the one term whose optimum depends on the whole team's bodies, since every agent must
bend by an amount set by its own and its partners' height and reach; it is the step that fails without body
descriptions (\cref{tab:abl_cond}).

\subsection{Task-Specific Interaction Rewards}
A skill keeps the reward of the last phase from its first epoch and changes only what its task requires
(\cref{tab:skill_rewards}). Along a path, transport, alignment and progress are paid towards the current waypoint, and
the next waypoint becomes the target once the table is within 0.4\,m of it, so that the team is paid for moving on; the path term keeps the table close to the segment between waypoints, and the
put-down mode latches only at the last waypoint. The pose skill keeps the path terms and adds a target heading of the table at every waypoint:
the heading term pays the error to it while the team grips, and the team moves to the next waypoint, or puts the table
down at the last, only once the error is below $30^\circ$, loosened by $3^\circ$ for every agent beyond two. On stairs,
every height entering the hand, contact, lift and put-down terms is measured above the ground below that point, so
that the carry and put-down heights follow the steps. A level table over a step stands higher above the lower tread
than above the upper one, so the lift term drops its penalty above $z^*$; progress counts twice. A box has no rim to
share: each agent is assigned one of the box's grasp points, walks to a standing point beyond it, holds both hands at it,
lifts it to 0.6\,m and carries the box to the target, where the box is aligned and set down. This reward extends CooHOI's
two-agent reward~\cite{gao2024coohoi} to teams of two to four, with a carry-speed term and the signed progress term added; the box skill is scored
by a single discriminator on CooHOI's motion clips, without the arm mask.

\section{Training}
\label{sec:app_training}

\begin{figure}[t]
  \centering
  \includegraphics[width=\columnwidth]{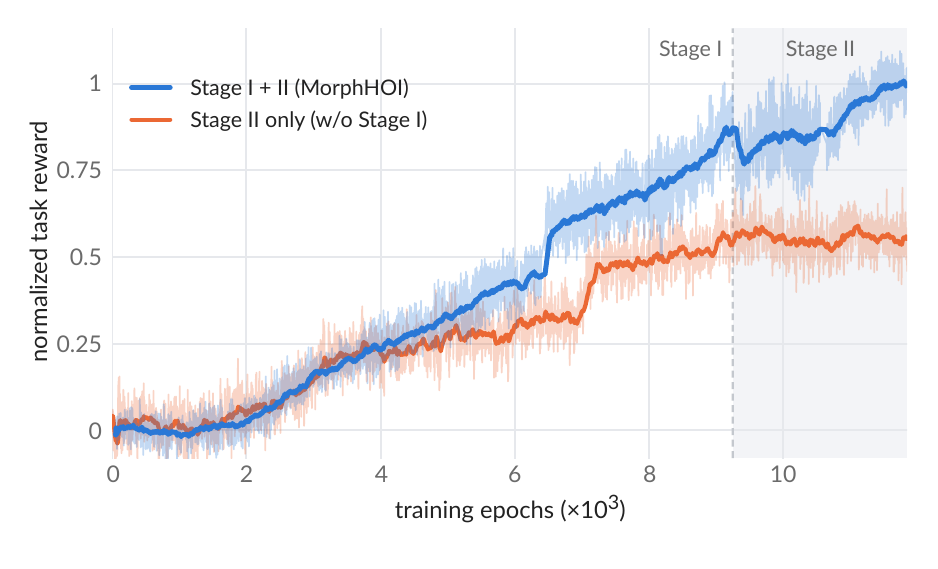}
  \vspace{-1.8em}
  \caption{\textbf{Training task reward} of the full recipe and of stage II alone,
  trained from the start on mixed teams, normalised to the initial policy (0) and the end of the full recipe (1). Both
  runs follow the two reward phases of \cref{tab:rewards}; the steps near epochs 6400 and 6900 mark their switch to the
  second phase.}
  \label{fig:reward}
\end{figure}

\subsection{Episodes and Teams}
At each reset, a team size $k \sim \mathcal{U}\{2,3,4\}$ is sampled, exposing the shared policy to varying numbers of cooperating agents throughout training. Advantages are normalised separately for each team size over the active agents, preventing differences in team size from affecting the scale of policy updates. Agents spawn up to 8\,m from the table, at least 1\,m apart, and the target
lies 3-10\,m from the table. An agent falls when a body part other than the feet touches the ground below a height
set by its leg length; any fall, a dropped table or a table lifted too high ends the episode for the whole team. The
team composition of an environment is fixed for the run: in stage I each environment draws a team of identical bodies,
with the body drawn uniformly over the training bodies, and in stage II 70\% of the environments fill each slot
independently and uniformly over the training bodies while the others keep teams of identical bodies.

\subsection{Training Schedule}
\method is trained from scratch with PPO on teams of identical bodies in stage I. Stage I ends at epoch 9250, and stage II adds 2600 epochs. Training runs in Isaac Gym~\cite{makoviychuk2021isaac} with PD control at 30\,Hz and PPO~\cite{schulman2017proximal} with GAE~\cite{schulman2015high} (\cref{tab:train}), with 256 environments per GPU on four GPUs. The reward follows two phases (\cref{tab:rewards}). In the first, progress is rewarded at half weight, and episodes last 600 steps, allowing agents to learn gripping, lifting, and carrying without penalising unstable movements early in training. Once the reward plateaus, the second phase introduces signed progress at full weight to penalise movement away from the target, adds stay and lowering rewards activated upon arrival, and extends episodes to 750 steps to allow time for putting the table down. \Cref{fig:reward} compares the task reward of this recipe with training stage II alone from scratch on mixed teams, using the same two reward phases. Each run switches phase once its first-phase reward plateaus, at epoch 6400 for the full recipe and 6900 for stage II alone. The curves rise similarly for the first 4000 epochs, after which stage II alone falls behind and plateaus near half the final reward of the full recipe. Training directly on mixed teams makes early coordination more difficult, resulting in fewer successful deliveries and infrequent arrival-based rewards. This limits learning beyond partial carrying behaviours, preventing the policy from reliably completing the full task. In contrast, stage I establishes basic cooperation among identical bodies before introducing heterogeneous teams. At the start of stage II, the full recipe shows a temporary reward drop as mixed teams replace identical ones, followed by recovery as the policy adapts to heterogeneous cooperation.

\subsection{Skill Adaptation}
Each skill adds one tokenizer and the gated low-rank updates of \cref{eq:lora} to the frozen stage-II policy, and each
new tokenizer first trains alone for 200 epochs. The path skill draws five waypoints spaced 1.2-1.8\,m apart with
turns of up to $\pm 40^\circ$ in 80\% of the episodes and observes the next three waypoints; teams of two to four train
for 2600 epochs on episodes of 900 steps. The pose skill follows the same waypoints and, like every other skill, adds its
own tokenizer and gated updates directly to the frozen stage-II policy; in 64\% of the episodes each waypoint carries a target heading, the path tangent or a random walk of up to
$\pm 45^\circ$ per waypoint within $\pm 120^\circ$, scaled in over 800 epochs, and teams of two to four train for
3000 epochs on episodes of 1000 steps. The stairs skill observes a $12 \times 7$ grid of ground heights with a spacing of 0.13\,m
around each agent and eight height samples every 0.5\,m along the line from the table to the target (two tokens);
teams of two to four train on four steps of 2-12\,cm on episodes of 1200 steps, for 7000 epochs downwards and 7000 upwards. The box
skill observes the box's grasp points (one token), uses rank 32 instead of 8 and 16, and trains for 10000 epochs on teams of two to four.

\subsection{Baselines}
Both retrained baselines learn a single policy for the training bodies of \method, and use the same total budget of 11{,}850 epochs as stages I and II of \method. Neither
receives a body description, a contact assignment or the relational bias: one set of weights must control every body
from its canonical observations alone. Because the bodies' joint spaces differ, the original methods cannot be applied
to them directly; both therefore observe and act through the canonical interface of \cref{sec:canonical} and are
scored by its stature-normalised discriminator, but keep the rewards of their original settings, without the
per-body style weight of \cref{eq:reward} or the body-dependent contact gate of \cref{eq:gate}. TeamHOI$^*$ keeps
TeamHOI's network and its original reward,
walking and lifting until epoch 3400 and the full reward afterwards. CooHOI$^*$ is initialised as the CooHOI baseline of TeamHOI~\cite{lionar2026teamhoi},
which adapts CooHOI~\cite{gao2024coohoi} to teams, and it trains with the
reward that TeamHOI describes for this baseline.

%% file: tables/appendix_table3_rewards.tex
\begin{tabularx}{\textwidth}{@{}l l >{\raggedright\arraybackslash}X c c c@{}}
\toprule
 & Term & Functional form & I$_{1\text{--phase 1}}$ & I$_{\text{phase 1--phase 2}}$ & II \\
\midrule
\multirow{5}{*}{Approach} & walk\_pos  & $\exp\!\big(-2\,(d_\mathrm{nn} - 0.3)^2\big)$ & 0.2 & 0.2 & 0.2 \\
 & walk\_vel  & $\mathbf{1}[v_n > 0]\,\exp\!\big(-2\,(v_n - \mathrm{clip}(v_n, 1.5, 2.5))^2\big)$ & 0.4 & 0.4 & 0.4 \\
 & walk\_face & $\big(r_\mathrm{ang}\,[(1-\sigma_d)\,[n^\top\! f]_+ + \sigma_d\,[u_c^\top\! f]_+]\big)^{1/2}$ & 0.2 & 0.2 & 0.2 \\
 & formation  & $0.75\, r_\mathrm{cov} + 0.25\, r_\mathrm{ang}$ & 0.6 & 0.6 & 0.6 \\
 & hand       & $r_\mathrm{cov}\, \mathbf{1}[d_\mathrm{nn} \le 1]\, \mathrm{mean}_h\, \phi_h \exp(-5\, d_h)$ & 0.7 & 0.7 & 0.7 \\
\midrule
\multirow{2}{*}{Grip} & contact    & $\min_h\, [1 - d_h / a_e]_+$ & 0.7 & 0.7 & 0.7 \\
 & lift       & $r_\mathrm{cov}\, \tfrac12 \textstyle\sum_{h \in \mathcal{C}} \rho_h \exp(-5\,[z_h - z^*]_+)$ & 1.4 & 1.4 & 1.4 \\
\midrule
\multirow{3}{*}{Carry} & transport  & $G \exp(-0.15\, \lVert x_g - x_o \rVert^2)$ & 1.0 & 1.0 & 1.0 \\
 & align      & $G\, [u_g^\top\! f_\mathrm{far}]_+$ & 0.4 & 0.4 & 0.4 \\
 & progress   & $G\, \mathrm{clip}(\dot d / 0.6,\, -1,\, 1)$ & 0.5 & 1.0 & 1.0 \\
\midrule
\multirow{3}{*}{Arrival} & putdown    & $P\, \big(0.8\, \min_h \exp(-5\, |z_h - z_\mathrm{put}|) + 0.2 \exp(-2\, \lVert v_\mathrm{root} \rVert)\big)$ & 1.0 & 1.0 & 1.0 \\
 & stay       & $P \exp(-5\, \lVert x_g - x_o \rVert)$ & \off & 1.0 & 1.0 \\
 & lower      & $P \exp(-10\, [z_o - z_o^\mathrm{rest}]_+)$ & \off & 1.0 & 1.0 \\
\midrule
Style & style & $- \log \max(1 - D,\, 10^{-4})$ & 0.3 & 0.3 & 0.3 \\
\midrule
\multicolumn{3}{@{}l}{Episode length (steps)} & 600 & 750 & 750 \\
\bottomrule
\end{tabularx}

%% file: tables/appendix_table4_skill_rewards.tex
\begin{tabular}{@{}l l l c@{}}
\toprule
Skill & Term & Functional form or change & $w$ \\
\midrule
\multirow{3}{*}{Path} & path    & $G \exp(-4\, e_\mathrm{seg}^2)$ & 0.5 \\
 & target  & $x_g \leftarrow x_j$;\ $j \leftarrow j{+}1$ if $\lVert x_j - x_o \rVert < 0.4$ & -- \\
 & $P$     & latched at the last waypoint only & -- \\
\midrule
\multirow{2}{*}{Pose} & heading & $G \exp(-5\, e_\psi^2)$ & 0.6 \\
 & $j$, $P$ & also need $e_\psi < 30^\circ + 3^\circ (k - 2)$ & -- \\
\midrule
\multirow{3}{*}{Stairs} & heights & $z \leftarrow z - z_\mathrm{gnd}(x, y)$ & -- \\
 & lift    & $\tfrac12 \textstyle\sum_{h \in \mathcal{C}} \rho_h$ & 1.4 \\
 & progress & $G\, \mathrm{clip}(\dot d / 0.6,\, -1,\, 1)$ & 2.0 \\
\midrule
\multirow{10}{*}{Box} & walk\_pos   & $\exp(-2\, \lVert p_s - p_\mathrm{root} \rVert^2)$ & 0.1 \\
 & walk\_face  & $[u_p^\top\! f]_+$ & 0.1 \\
 & hold\_hand  & $\exp(-5\, \lVert \bar p_\mathrm{hand} - p_g \rVert^2)$ & 0.3 \\
 & hold\_height & $\exp(-10\, (0.6 - z_g)^2)$ & 0.1 \\
 & carry\_far  & $L \exp(-0.5\, \lVert x_g - x_o \rVert^2)$ & 0.3 \\
 & carry\_near & $L \exp(-10\, \lVert \mathbf{x}_g - \mathbf{x}_o \rVert^2)$ & 0.2 \\
 & carry\_vel  & $L\, \mathbf{1}[\dot d > 0] \exp(-2\, [1 - \dot d]_+^2)$ & 0.2 \\
 & progress    & $L\, \mathrm{clip}(\dot d / 0.6,\, -1,\, 1)$ & 1.0 \\
 & carry\_dir  & $N\, [a_o^\top a_g]_+$ & 0.1 \\
 & putdown     & $N \exp(-5\, (z_g - 0.25)^2)$ & 0.1 \\
\bottomrule
\end{tabular}

%% file: tables/appendix_table5_abl_ect.tex
\begin{tabular}{@{}l cccc c cccc@{}}
\toprule
 & \multicolumn{4}{c}{Seen bodies} && \multicolumn{4}{c}{Unseen embodiments} \\
\cmidrule{2-5}\cmidrule{7-10}
 & SR$\uparrow$ & DE$\downarrow$ & CR$\uparrow$ & SPARC$\uparrow$ && SR$\uparrow$ & DE$\downarrow$ & CR$\uparrow$ & SPARC$\uparrow$ \\
\midrule
Flat attention & 90.4 & 0.43 & 78.8 & $-$2.08 && 88.2 & \textbf{0.55} & 66.4 & $-$2.08 \\
w/o OT task bias & 84.2 & 0.71 & 77.5 & $-$1.99 && 80.0 & 0.99 & 64.9 & $-$2.03 \\
w/o relational bias & 87.4 & 0.57 & 78.2 & $-$2.01 && 82.1 & 0.84 & 64.7 & $-$2.06 \\
\method & \textbf{93.0} & \textbf{0.38} & \textbf{79.6} & \textbf{$-$1.92} && \textbf{89.8} & 0.56 & \textbf{67.4} & \textbf{$-$1.97} \\
\bottomrule
\end{tabular}

%% file: tables/appendix_table6_abl_iface.tex
\begin{tabular}{@{}l cccc c cccc@{}}
\toprule
 & \multicolumn{4}{c}{Seen bodies} && \multicolumn{4}{c}{Unseen embodiments} \\
\cmidrule{2-5}\cmidrule{7-10}
Action space & SR$\uparrow$ & DE$\downarrow$ & CR$\uparrow$ & SPARC$\uparrow$ && SR$\uparrow$ & DE$\downarrow$ & CR$\uparrow$ & SPARC$\uparrow$ \\
\midrule
Hinges by name & 35.1 & 3.07 & 67.3 & $-$2.55 && 31.0 & 3.25 & 61.1 & $-$2.93 \\
Links + IK & 82.0 & 0.88 & 74.5 & $-$1.99 && 78.1 & 0.92 & 64.0 & $-$2.17 \\
Joints + IK (\method) & \textbf{93.0} & \textbf{0.38} & \textbf{79.6} & \textbf{$-$1.92} && \textbf{89.8} & \textbf{0.56} & \textbf{67.4} & \textbf{$-$1.97} \\
\bottomrule
\end{tabular}

%% file: tables/appendix_table7_abl_zswap.tex
\begin{tabular}{@{}l cccc c cccc@{}}
\toprule
 & \multicolumn{4}{c}{Seen bodies} && \multicolumn{4}{c}{Unseen embodiments} \\
\cmidrule{2-5}\cmidrule{7-10}
Given $z_e$ & SR$\uparrow$ & DE$\downarrow$ & CR$\uparrow$ & SPARC$\uparrow$ && SR$\uparrow$ & DE$\downarrow$ & CR$\uparrow$ & SPARC$\uparrow$ \\
\midrule
\method & \textbf{93.0} & \textbf{0.38} & \textbf{79.6} & \textbf{$-$1.92} && 89.8 & 0.56 & \textbf{67.4} & \textbf{$-$1.97} \\
Teammate's & 71.2 & 1.49 & 58.4 & $-$2.01 && 68.2 & 1.69 & 49.1 & $-$2.08 \\
Farthest body's & 66.2 & 1.56 & 54.9 & $-$2.07 && 61.6 & 1.92 & 44.1 & $-$2.11 \\
\midrule
\multicolumn{10}{@{}l}{\textit{Own $z_e$ + Gaussian noise (std $\times$ training spread)}} \\
$\sigma = 0.1$ & 91.7 & 0.48 & 76.0 & $-$1.96 && 90.2 & 0.58 & 62.5 & $-$2.01 \\
$\sigma = 0.5$ & 90.9 & 0.47 & 74.6 & $-$1.96 && \textbf{90.3} & \textbf{0.50} & 63.3 & \textbf{$-$1.97} \\
$\sigma = 1.0$ & 89.6 & 0.50 & 69.2 & $-$2.02 && 86.1 & 0.65 & 58.0 & $-$2.03 \\
\bottomrule
\end{tabular}